\documentclass{article}

\usepackage[margin=1in]{geometry}
\usepackage{graphicx}
\graphicspath{{./}}
\usepackage{subcaption}
\usepackage{amsmath}
\usepackage{amssymb}
\usepackage{booktabs}
\usepackage{multirow}
\usepackage{xcolor}
\usepackage[round]{natbib}
\usepackage[colorlinks=true,linkcolor=blue,citecolor=blue,urlcolor=blue]{hyperref}

\newcommand{\mvh}{\mbox{membership head}}   
\newcommand{\bvh}{\mbox{Boolean head}}      
\DeclareMathOperator{\GELU}{GELU}
\newcommand{\sig}{\sigma}

\title{Legible-by-Construction: Attention and End-to-End Transformers}
\author{Mark Oskin\\
Professor\\
School of Computer Science and Engineering\\
University of Washington\\
\texttt{mhoskin@uw.edu}}
\date{July 2026}

\begin{document}

\maketitle

\begin{abstract}
A companion paper showed that a transformer's feed-forward layer can be rebuilt from explicit fuzzy
set operations---intersection, set-difference, and a self-forgetting sequence quantifier---so that its
hidden units read as named logical operators at no cost to language-model quality \citep{oskin2026ncffn}.
That left the other half of the transformer opaque. Here we carry the same idea into attention and then
join the two into a single model. The core mechanism is deliberately minimal: a head's \emph{value} is passed
through a sigmoid, $v=\sig(W_v x)\in[0,1]$, so each value channel becomes a readable ``does this feature
hold'' detector. This adds no parameters and leaves the standard head otherwise untouched---part of the
point, since so small a change should not buy this much legibility, but it does. A Boolean variant goes
further, restructuring the value into an explicit within-token intersection and negation-capable
set-difference for crisper, explicitly logical relations, at the cost of departing from the standard value
projection. In both designs the output projection is left \emph{free}---not tied to the vocabulary---which is
the load-bearing decision: bounding what a head \emph{detects} while leaving what it \emph{writes}
unconstrained yields selective detectors, whereas constraining the write instead does not. A bounded value is
then shaped into a readable detector by two \emph{selectivity pressures}: one that makes a channel fire on a
sparse set of contexts, one that makes it fire decisively at the rails rather than hovering in between. Which
pressure a design wants is not universal---a single membership value wants both, whereas an explicit
conjunction already fires sparsely by construction, so pressing its operands toward crispness \emph{alone}
keeps them alive and readable while the added sparsity pressure starves them into dead constants. Across five
specialized-attention designs at $125$M parameters, $44$--$62\%$ of value channels become crisp, contextually
selective detectors, and---unlike the folklore that ``attention heads crystallize only on punctuation''---their
legibility \emph{rises with depth}, so that the deepest layers read as recognizable semantic and grammatical
features rather than punctuation. Language-model quality is at parity with a conventional baseline. Finally, we
couple the Boolean attention to the legible feed-forward layer and train an \emph{end-to-end
legible-by-construction} language model: at baseline perplexity and downstream accuracy---the strongest
variant sitting at the top of the parity band on LAMBADA and BLiMP---its feed-forward units are named set and
quantifier operations throughout, and we can take a token it generates and read the named units that compose to
produce it. A sharp trainability boundary on which mixtures of bounded and unbounded heads train stably is
characterized in an appendix. We argue this changes the kind of object a language model \emph{is} to work
with---an object whose internal computation can be read, edited, and audited by construction rather than
recovered post hoc---and we lay out what such models make possible.
\end{abstract}

\section{Introduction}
\label{sec:intro}

Most of what we know about the inside of a transformer we know \emph{after the fact}. We train an opaque
model and then recover structure from it---probing directions, projecting activations through the
unembedding \citep{nostalgebraist2020logitlens,belrose2023tunedlens}, or fitting sparse dictionaries to
decompose superposed features \citep{bricken2023monosemanticity,cunningham2023sae}. This post-hoc program
has been remarkably productive, but it is a program of \emph{reconstruction}: the legibility lives in an
external tool, not in the model, and it is always partial, contestable, and expensive to maintain as the
model changes.

A different route is to make the model legible \emph{by construction}---to choose architectural primitives
whose computation is a named operation before any interpretation is attempted. A companion paper took this
route for the feed-forward network (FFN), the two-thirds of a transformer's parameters that sit outside
attention \citep{oskin2026ncffn}. It replaced the FFN's opaque activation with explicit fuzzy set
operations: each hidden unit computes an intersection $A\cap B$ or a negation-capable set-difference
$A\setminus B$ over sigmoid-bounded operands, and a self-forgetting sequence quantifier supplies the one
temporal primitive a left-to-right model needs. The result was a feed-forward layer whose units are named
logical operations, a readable subset of which act as grammatical-licensing detectors, at no cost to
language-model perplexity. But that work left the other half of the transformer---attention---exactly as
opaque as before. A model with a legible FFN and an illegible attention is still, as an object, illegible.

This paper closes that gap and then joins the halves. Our starting observation is an old asymmetry in the
attention circuit \citep{elhage2021framework}: the query--key (QK) product is comparatively legible---it is
a pattern over positions, a readable \emph{where}---while the output--value (OV) path is the opaque
\emph{what}. Reading a head's value by projecting it to the vocabulary fails; the folklore
conclusion, borne out in practice, is that attention heads rarely ``crystallize'' into anything nameable
beyond a few that track punctuation or position. We show this folklore is a statement about the
\emph{basis} one reads in, not about attention. The fix is the same one that worked for the FFN, transposed
to the value, and it is deliberately minimal: bound the head's value into fuzzy memberships,
$v=\sig(W_v x)\in[0,1]$, so that each value channel is a readable ``does this feature hold at this token''
detector, and let attention fuzzy-aggregate those memberships over context. This \emph{membership} head adds
no parameters and changes nothing else about a standard head---only a sigmoid on the value, plus a light
selectivity pressure that keeps the detectors sparse and crisp---so it is striking that so small a change makes the value readable at
all. A \emph{Boolean} variant goes one step further, restructuring the value into an explicit within-token
intersection and set-difference (the OV analogue of the FFN's set operators) for crisper logical relations, at
the price of departing from the standard single value projection.

The single most important design decision is \emph{where} the constraint goes. We constrain what the head
\emph{detects} (its bounded value, pre-output-projection) and leave what it \emph{writes} (the output
projection) completely free. Constraining the \emph{write} instead---for instance tying a head's output to
the vocabulary so it decodes to tokens by construction---does not work, and Section~\ref{sec:architecture}
explains why; the short version is that it opens a gradient shortcut a context-independent constant can
satisfy. This bounded-value / free-readout principle is what makes attention legible without making it worse, and
it is the conceptual core of the paper.

We report four things.
\begin{itemize}
\item \textbf{Legible attention heads (Section~\ref{sec:results-attn}).} Across five specialized-attention designs
at $125$M parameters, $44$--$62\%$ of value channels become crisp, contextually selective detectors, versus
the constant-collapse of the vocabulary-tied design. Crucially, their legibility \emph{increases with
depth}: the deepest layers reach $70$--$95\%$ selective and read as recognizable semantic and grammatical
features (an abstract-concept channel, a domain channel, a people/plural channel), directly contradicting
the ``heads only crystallize on punctuation'' folklore.
\item \textbf{Which pressure fits which design (Sections~\ref{sec:selectivity},~\ref{sec:results-attn}).} A
bounded value is made selective by two training pressures---one toward firing on a \emph{sparse} set of
contexts, one toward firing \emph{crisply} at the rails---and the right combination is not universal. A single
membership value wants both. An explicit conjunction already fires sparsely by construction, so it wants
crispness \emph{alone}: adding the sparsity pressure over-sparsifies its operators into dead constants. And
the pressures must target the \emph{operands} of an operator, not its combined output, on which they are
inert. Which pressures a design wants is a property of the operator it implements, not a global knob.
\item \textbf{Parity (Section~\ref{sec:results-attn}).} On LAMBADA, BLiMP, and ARC-Easy the designs sit at
parity with a conventional baseline, differences within what a single seed can resolve. The one real cost is a
LAMBADA-perplexity gap for the fully-bounded membership design---largest under sparsity alone, and recovering
under both pressures.
\item \textbf{An end-to-end legible language model (Section~\ref{sec:results-e2e}).} Coupling the Boolean
attention to the legible FFN yields a $125$M model at baseline perplexity and downstream accuracy---the best
variant at the top of the parity band on LAMBADA and BLiMP---whose feed-forward units are named set and quantifier
operations throughout. We can then take a token the model generates and decompose its logit into the named
units that produce it: for ``it was the best of times, it was the~\_\_'' the model writes ``worst,'' and the
attribution reads as a superlative detector intersected with a negation detector---the computation
read directly in its own basis rather than reconstructed after the fact.
\end{itemize}

The attention legibility is a \emph{deep-layer} phenomenon---shallow
attention, once coupled to a legible FFN, goes constant---and our language-model numbers are single-seed at
one scale. Legibility also comes with a training constraint: certain mixtures of a bounded majority of heads
with a small unbounded minority diverge reproducibly, a boundary we characterize and trace to a
saturated-head runaway write in Appendix~\ref{sec:trainability}. None of these undercuts the central claim,
which is architectural: a transformer can be built so that
its dominant computations are named operations, readable and editable without a post-hoc tool, at no quality
cost that we can measure. Section~\ref{sec:discussion} argues why that is worth having and what it makes
possible.

\section{Background}
\label{sec:related}

Our contribution sits at the intersection of two literatures that have stayed separate: mechanistic
interpretability of attention, which is overwhelmingly \emph{post-hoc}, and interpretable-by-construction
architecture, which has concentrated on the feed-forward layer and the classification setting. We organize
the comparison around two axes that locate our design precisely: \emph{pattern vs.\ value} (does the
constraint act on the QK attention map or the OV value?) and \emph{pinned vs.\ free readout} (is the
interpretable quantity forced into the vocabulary basis, or read in its own basis?).

\begin{figure}[t]
\centering
\includegraphics[width=0.98\textwidth]{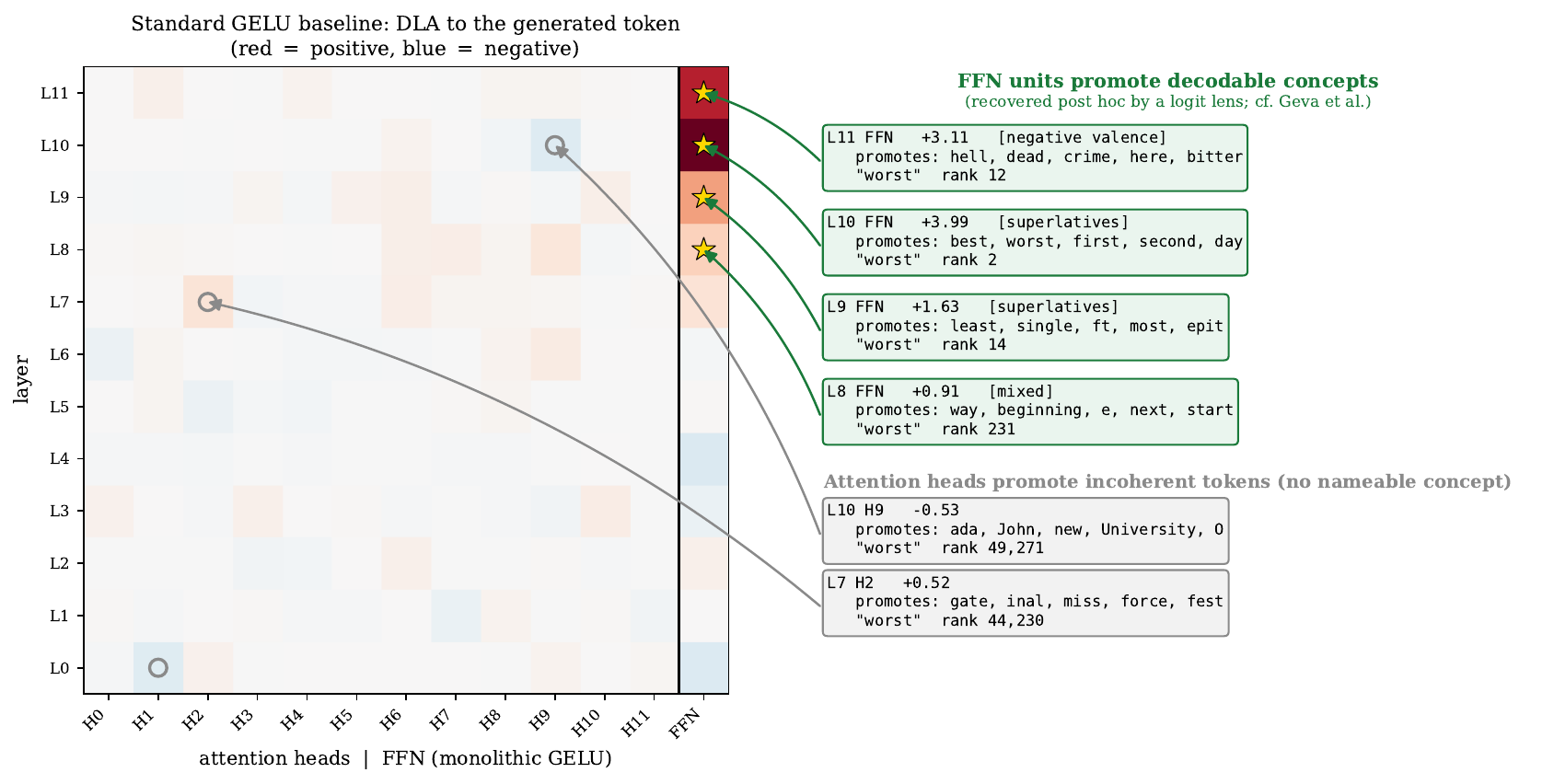}
\caption{Legibility in a \emph{standard} transformer is real, but post-hoc and one-sided---the situation this
paper changes. Direct logit attribution of the token a conventional \textsc{gelu} baseline generates for
``It was the best of times, it was the'' (again \emph{``worst''}), across all $12$ layers and per attention
head / feed-forward block. The feed-forward writes decode, through a logit lens, into coherent vocabulary
concepts: a \textsc{superlative} unit (layer~$10$, \emph{best}/\emph{worst}/\emph{first}, ``worst'' at rank
$2$) and a \textsc{negative-valence} unit (layer~$11$, \emph{hell}/\emph{dead}/\emph{crime}) compose to
``worst,'' reproducing the vocabulary-space view of feed-forward layers \citep{geva2021kv,geva2022promote}.
The attention heads, by contrast, promote incoherent tokens (``worst'' beyond rank $44{,}000$): the value is
opaque. Both readings exist only \emph{after} training, via a lens. Our model instead names the feed-forward
operators \emph{by construction} and makes the attention value legible as well (Figure~\ref{fig:dla}).}
\label{fig:dla-baseline}
\end{figure}

\paragraph{Reading attention after the fact.}
The dominant approach recovers structure from a trained model. The logit lens projects intermediate
activations through the unembedding \citep{nostalgebraist2020logitlens}, and the tuned lens learns a
per-layer affine correction because the raw projection reads as noise in the middle of the network
\citep{belrose2023tunedlens}---an observation we reproduce and that motivates \emph{not} reading values in
the vocabulary basis at all. Vocabulary-space analyses of transformer components
\citep{geva2021kv,geva2022promote,dar2023analyzing} decode feed-forward and attention writes into token
distributions; Attention Lens learns per-head transforms to make head outputs decodable
\citep{sakarvadia2023attnlens}; and attention-output sparse autoencoders decompose the OV stream into
features after training \citep{kissane2024attnsae}. All of these are reconstructions of a fixed opaque model, and
Figure~\ref{fig:dla-baseline} makes concrete what there is to reconstruct: in a conventional baseline the
feed-forward writes already decode into coherent concepts---a \textsc{superlative} unit and a
\textsc{negative-valence} unit compose to produce the generated token, reproducing
\citet{geva2021kv,geva2022promote}---while the attention value stays opaque, and the whole picture is available
only \emph{after} training, through a lens.
We instead make the value legible in the model, so that no lens or dictionary is needed to state what a head
detects.

 Recent work also finds that many genuine features are \emph{dense}, not sparse, and are not
vocabulary-aligned \citep{sun2025denselatents}; this is consistent with our own finding that a conventional
value does not decode to tokens, and is why we constrain the value directly rather than hoping it decomposes.

\paragraph{The QK/OV legibility asymmetry.}
The circuits framework separates a head into a QK circuit that decides \emph{where} to attend and an OV
circuit that decides \emph{what} to move \citep{elhage2021framework}. The QK pattern is comparatively
readable; the OV value is the opaque half. Almost all subsequent interpretability of attention has read the
pattern. Our design targets the neglected half---the value---and this is the axis on which we differ from
nearly all prior structured-attention work.

\paragraph{Bounded, sigmoid, and fuzzy attention---on the pattern.}
A growing line replaces or bounds the softmax to make the attention \emph{map} better-behaved or more
interpretable: sigmoid attention removes the row-normalization coupling \citep{ramapuram2024sigmoidattn},
and an attention \emph{bottleneck} bounds the total attention mass for user intervention and debugging
\citep{rahmanzadehgervi2024tab}. Concept Transformers make a classifier's attention \emph{pattern} range over a
supervised bank of named concepts, with the readout pinned to those concepts \citep{rigotti2022concept}.
Every one of these acts on the pattern (the \emph{where}); none bounds the value-as-membership (the
\emph{what}), and the interpretable quantity is generally a supervised or pinned target. Ours is the
value-side, unsupervised, free-readout counterpart: the head's memberships are discovered from the
language-modeling objective, not named in advance, and are read in their own basis.

\paragraph{Interpretable-by-construction units.}
On the feed-forward side, Softmax Linear Units sharpen neuron activations toward monosemanticity
\citep{elhage2022solu}; codebook features quantize activations into a discrete, inspectable code
\citep{tamkin2023codebook}; and bilinear MLPs expose weight-based structure by removing the elementwise
nonlinearity \citep{pearce2024bilinear}. These make the FFN more analyzable but do not give each unit a
\emph{named logical} form, and they do not address attention. Our companion paper's feed-forward layer
\citep{oskin2026ncffn}---explicit fuzzy set operations with a self-forgetting quantifier---does give the
named form; the present paper supplies the missing attention piece and shows the two compose into a single
legible model. The distinction from codebook/discrete methods is that our value is a \emph{fuzzy} membership
in $[0,1]$ with a free continuous readout, not a discrete code with a fixed decoder.

\paragraph{Vocabulary-tied readouts, and why we avoid them.}
The natural way to make a head ``decodable by construction'' is to tie its output to the token embedding, so
its write is a token distribution. We built exactly this and it failed: the value collapsed to a
context-independent constant, because pinning the output to the vocabulary opens a gradient path to the
unigram prior that a constant satisfies. This is the concrete lesson behind our pattern of constraint---bound
the \emph{detected feature} (value, pre-projection) and leave the \emph{write} (output projection) free---and
it is the axis (pinned vs.\ free readout) on which we differ from Concept Transformers and from
vocabulary-space methods. It also connects to superposition: because genuine value content is dense and
non-lexical \citep{elhage2022superposition,sun2025denselatents}, forcing it into the vocabulary basis is
actively harmful, whereas a bounded membership with a free readout keeps the content in a basis it can use.

\paragraph{Fuzzy and differentiable logic.}
The set operations we use are standard product-t-norm fuzzy connectives \citep{vankrieken2022fuzzy}, in the
tradition of logic tensor networks \citep{badreddine2022ltn} and differentiable logic gate networks
\citep{petersen2022difflogic}, and the multiplicative form relates to gated and bilinear FFNs
\citep{dauphin2017glu,shazeer2020glu,pearce2024bilinear}. Our use is distinctive in being a drop-in,
parameter-neutral component of a \emph{language model} whose legibility is measured on real text, and in
being applied to the attention value rather than to a dedicated reasoning module. The self-forgetting
quantifier's learned decay is the classical forget gate \citep{gers2000forget} repurposed as a legible
temporal operator, detailed in the companion paper.

\paragraph{Superposition and automated interpretation.}
We still find polysemantic channels, and we quantify legibility in the vocabulary of that literature:
crisp-but-globally-polysemantic units whose meaning is local to an input \citep{elhage2022superposition},
the monosemanticity goal of dictionary learning
\citep{bricken2023monosemanticity,cunningham2023sae}, and the name-then-predict evaluation of automated
interpretability \citep{bills2023autointerp}. The difference is that our units come with a named
\emph{operation} for free; superposition then governs only whether the \emph{operands} are monosemantic, a
strictly smaller problem.

\paragraph{Training instabilities.}
The trainability boundary we characterize (Appendix~\ref{sec:trainability}) is a structured, reproducible
instance of the late-onset divergences studied at scale
\citep{wortsman2023smallscale}; our contribution is not the observation that transformers can diverge but a
specific, checkpoint-verified mechanism---saturating bounded heads plus a starved unbounded minority---that
predicts \emph{which} mixtures diverge and why the relationship is non-monotonic in the bounded fraction.

\section{Architecture}
\label{sec:architecture}

Our method changes one object inside a standard multi-head attention layer---the value---and leaves everything
else untouched: the queries, the keys, the attention map, the output projection, and the residual stream are
all exactly as in a conventional transformer. The change is governed by a single principle, which we state
first because it---and not the particular bounded form the value takes---is what the paper is about:
\emph{constrain what a head detects, and leave what it writes free}. The rest of this section makes that
principle precise (\S\ref{sec:arch-circuits}--\S\ref{sec:arch-where}) and gives two ways to instantiate it: a
\emph{minimal} one---a fuzzy \emph{membership} value, just a sigmoid on an otherwise standard head---and a
\emph{structured} one---an explicit \emph{Boolean} value that restructures the head's value projection into
set operations. It then describes how we mix bounded heads with conventional ones, how we pressure them to be selective, and how we couple the result to the
legible feed-forward layer of \citet{oskin2026ncffn}. Every design is parameter-neutral relative to its
conventional baseline: the membership head reuses the value projection unchanged, and the Boolean head splits
it into two half-width projections of equal total size, so neither introduces new weights (the sigmoids and
products add none).

\subsection{Two circuits, one opaque}
\label{sec:arch-circuits}
For input $X\in\mathbb{R}^{T\times d}$ and head $h$ with head dimension $d_h$, a standard head computes a
value $V_h = X W_v^h$, an attention map $A_h=\mathrm{softmax}(Q_hK_h^\top/\sqrt{d_h})$, and an output
$O_h = A_h V_h$; the per-head outputs are concatenated and mixed by a shared output projection $W_O$
\citep{vaswani2017attention}. The circuits view \citep{elhage2021framework} splits this into two parts that
are usually studied separately. The \emph{QK circuit}---the attention map $A_h$---decides \emph{where} the
head reads, and it is already legible: an attention pattern can be plotted and read directly. The \emph{OV
circuit}---the value $V_h$ carried through $W_O$---decides \emph{what} the head moves from those positions
into the residual stream, and it is the opaque half. $V_h$ is a dense, unconstrained vector with no reason for
any of its coordinates to correspond to something a human can name; this is the part of attention that
post-hoc interpretability spends its effort trying to reconstruct. Our target is to make it legible
\emph{by construction} instead.

\subsection{Where to put the constraint}
\label{sec:arch-where}
Making the OV circuit legible means placing a constraint on it, and the central design choice---the one the
rest of the paper turns on---is \emph{which part} of the circuit to constrain. There are two candidates: the
value $V_h$, which is what the head detects \emph{before} $W_O$, or the output written to the residual stream,
which is what the head produces \emph{after} $W_O$. The two choices pull in opposite directions, and only one
of them works.

Constraining the \emph{output} is the more obvious choice, and it is the one we tried first. Tie a head's
post-$W_O$ write to the token-embedding matrix, so that whatever it adds to the residual stream decodes
directly to vocabulary tokens; then the write is legible by definition---one can just read off the tokens it
promotes. It fails, and instructively. Handed a gradient path that runs straight from the write to the
vocabulary, the head takes the cheapest available route to lower loss: it reproduces the unigram prior, and a
single context-independent constant satisfies that perfectly. The value stops varying with the input---it
detects nothing---and collapses to a bias. Legibility bought by pinning the write turns out to be the
legibility of a dead unit.

Constraining the \emph{value} instead leaves the write free and removes that shortcut. We give $V_h$ a
bounded, named form---defined in the next two subsections---so that each of its coordinates has a fixed
meaning, and we let $W_O$ learn, unconstrained, how to express those detected features in the residual
stream. The head is then read on the \emph{detection} side---which bounded features fired, and where the
attention pattern gathered them from---never by decoding its write. This is the principle stated at the top of
the section, and the collapse above is why the pre-$W_O$ side is the right place for the constraint: the value
is the half that must stay meaningful, and the write is the half that must stay free. Everything that follows
is a choice of what ``a bounded, named value'' should be.

\subsection{Membership value head}
\label{sec:membership}
The first instantiation makes each value coordinate a \emph{fuzzy membership}: a soft truth value in $[0,1]$
answering ``is feature $c$ present at this token?'', where $1$ is clearly yes, $0$ is clearly no, and
intermediate values are a partial or uncertain degree. The contrast with a conventional value is the point:
an unbounded value coordinate has no canonical scale or zero and means nothing on its own---only in concert
with $W_O$ and the rest of the network---whereas a membership coordinate carries a fixed, self-contained
reading on a common $[0,1]$ scale. Concretely, the \mvh{} passes the ordinary value through a sigmoid:
\begin{equation}
V_h \;=\; \sig\!\left(X W_v^h\right)\in[0,1]^{T\times d_h},
\qquad
O_h \;=\; A_h V_h .
\label{eq:membership}
\end{equation}
The head's trained parameters are exactly a conventional head's---the same per-head query, key, and value
projections $W_q^h, W_k^h, W_v^h$ and the shared output projection $W_O$---with a single sigmoid inserted on
the value. The membership head is thus parameter-identical to its baseline; its one difference is
functional---the sigmoid bounds each value coordinate to $[0,1]$, turning an unbounded activation into the
membership degree just described. Entry $(t,c)$ of $V_h$ is now the degree to which feature $c$ holds at
token $t$. Attention forms a
fuzzy-weighted average of these memberships over the attended context---so $O_h$ reports, per feature, how
strongly it was present among the positions the head read---and the free $W_O$ decides how those aggregated
memberships enter the residual stream. The bound to $[0,1]$ is not mere squashing, and it does two things. It endows each coordinate with a stable
semantics---a membership degree on a common scale---and therefore with a name, which an unbounded value, having
no shared frame, cannot supply. And it \emph{anchors the value's scale}: because the write $W_O$ is left free,
a sparsity pressure applied to an unbounded value would be rescaled away, whereas on a bounded membership it
cannot be, so the bound is precisely what lets the sparsity pressure of \S\ref{sec:selectivity} carve out a
sparse detector rather than a diffuse gate (\S\ref{sec:selectivity} makes this argument precise). Both roles
turn on constraining the value and not the write: this is the very same sigmoid bound that collapsed in
\S\ref{sec:arch-where} when we applied it after $W_O$ instead of before.
Reading the head means naming each channel by the contexts that drive it toward $1$ and inspecting its
attention pattern---never by decoding $O_h$ to tokens.

\subsection{Boolean value head}
The second instantiation goes one step further, and it is a larger architectural change. Where the membership
head left a standard head intact and merely bounded its value, the Boolean head alters the \emph{structure} of
the value: instead of leaving each detected feature to stand alone, it \emph{combines} two of them with an
explicit logical operation, making the value the OV analogue of the set operators in the legible FFN. Here the
single value projection is replaced by two half-width value
projections $W_a^h, W_b^h\in\mathbb{R}^{d\times d_h/2}$ (the per-head query and key projections and the shared
$W_O$ are unchanged); the head's trained parameters are now $W_q^h, W_k^h, W_a^h, W_b^h$ in place of the
conventional $W_q^h, W_k^h, W_v^h$. From these it detects two bounded operand banks $A=\sig(X W_a^h)$ and
$B=\sig(X W_b^h)$ with $A,B\in[0,1]^{T\times d_h/2}$, and forms the value as the concatenation of a fuzzy
intersection and a negation-capable set-difference:
\begin{equation}
V_h \;=\; \big[\,\underbrace{A\odot B}_{\text{intersection}}\;;\;\underbrace{A\odot(1-B)}_{\text{set-difference}}\,\big]\in[0,1]^{T\times d_h}.
\label{eq:boolean}
\end{equation}
A coordinate of the first block reads as ``$A$ \textsc{and} $B$'' (both features present), and one of the
second block as ``$A$ \textsc{and-not} $B$'' ($A$ present, $B$ absent); the two blocks are asymmetric by
construction, since $A\setminus B\neq B\setminus A$. Because $W_a^h$ and $W_b^h$ together have exactly the
size of the value projection $W_v^h$ they replace, the head stays parameter-neutral even though its value is
now produced by two trained projections rather than one. This is deliberately a \emph{within-token}
combination---a single attended token is simultaneously $A$ and not-$B$---rather than a cross-position
relation between argument slots. Relational binding across positions~\citep{smolensky1990tpr} is a different
mechanism that we do not attempt here, and one that, in our prior experiments, does not train profitably at
this scale.

\subsection{Fraction sweep}
We do not assume every head should be bounded. In a layer with $n_h$ heads we make the first $n_{\mathrm{sp}}$
special (bounded, via Eq.~\ref{eq:membership} or Eq.~\ref{eq:boolean}) and leave the remaining
$n_h-n_{\mathrm{sp}}$ conventional. The special \emph{fraction} $n_{\mathrm{sp}}/n_h\in\{50\%,75\%,100\%\}$
is swept for each family. As Appendix~\ref{sec:trainability} shows, this fraction is not a benign
hyperparameter: it controls a trainability boundary, and the relationship is non-monotonic.

\subsection{Selectivity pressures}
\label{sec:selectivity}
A bounded value is not automatically legible. To be readable a detector should fire on a \emph{specific,
sparse} set of contexts---so a human can name what it detects---and it should fire \emph{decisively}, near
$0$ or near $1$ rather than hovering in the ambiguous middle---so ``present'' and ``absent'' are unambiguous.
These are two separate properties, and we impose them with two separate \emph{selectivity pressures}, each a
small penalty on the bounded operands added to the language-model loss. By ``operands'' we mean the membership
value $\sig(XW_v^h)$ for a membership head, or the two operands $A,B$ for a Boolean head---the atoms we want
to name, not the combined output $A\cap B$ (a distinction Section~\ref{sec:results-attn} shows is essential).

\paragraph{Sparsity pressure.} A term $\lambda_{\mathrm s}\sum_{t,c}|v_{t,c}|$ pushes every value toward $0$,
so a channel is off by default and fires only where it earns its keep. Its effect depends on the bound. With
the output projection $W_O$ left free, a sparsity penalty on an \emph{unbounded} value is scale-defeated: the
network shrinks $v\!\to\!\alpha v$ and absorbs the factor into $W_O\!\to\!W_O/\alpha$, leaving the head's
function unchanged while driving the penalty toward zero, so it buys no sparsity. On a bounded value the
escape is closed---a coordinate of $\sig(\cdot)$ cannot be scaled toward $0$ without sending its
pre-activation to $-\infty$, i.e.\ genuinely turning the feature \emph{off}---so the pressure can be satisfied
only by real sparsity. The bound is what gives it teeth. (It is the static analogue of the legible FFN
quantifier's learned decay, and unlike the vocabulary-pinned design of \S\ref{sec:arch-where} it suffices on
its own.)

\paragraph{Crispness pressure.} A term $\lambda_{\mathrm c}\sum_{t,c} v_{t,c}(1-v_{t,c})$ is maximal at
$\tfrac12$ and zero at the rails, so it drives each value toward whichever rail is nearer---down when
$v<\tfrac12$, up when $v>\tfrac12$. Unlike sparsity it is \emph{symmetric} between $0$ and $1$: it makes a
value decisive without preferring off. It, too, cannot be rescaled away by the free $W_O$, because it
constrains the value's \emph{distribution} (pushing its mass to the rails), not its magnitude.

\paragraph{How they compose.} On each coordinate three forces act at once: the language-model loss pulls the
value \emph{up} toward $1$ wherever its presence---aggregated by attention or the FFN and expressed through
the free $W_O$---improves prediction; sparsity pulls \emph{down} toward $0$; crispness pulls toward the nearer
rail. Sparsity governs \emph{how many} coordinates fire, crispness \emph{how decisively} each sits, and the
task \emph{which} fire. Applied together they hold a value at $0$ unless the task lifts it past a threshold,
beyond which it is driven crisply to $1$---a sparse, crisp, contextual detector. The two can also be applied
separately, and \emph{which combination is right is not universal}: Section~\ref{sec:results-attn} shows that
for a single membership value sparsity is what makes it selective (both pressures together are best), whereas
for a conjunctive operator the intersection already supplies its own sparsity---$A\cap B$ fires only when both
operands do---so the crispness pressure \emph{alone} yields livelier, more readable operators, while adding
sparsity thins the operands until the operator rarely fires at all. Throughout we use
$\lambda_{\mathrm s}=10^{-3}$ and $\lambda_{\mathrm c}=3\times10^{-3}$.

\subsection{End-to-end legible model}
Finally we couple the Boolean attention (Eq.~\ref{eq:boolean}, at $100\%$) to the legible feed-forward layer
of \citet{oskin2026ncffn}, whose units are fuzzy set operations plus a self-forgetting sequence quantifier.
The FFN's operator fraction is set by a $\GELU$ partition (the $\rho$ of \citealp{oskin2026ncffn}): a
fraction of hidden units remain conventional $\GELU$ for training stability, the rest are the named
operators. We train two end-to-end configurations, differing only in that partition: a $25\%$-named split
($75\%$ of FFN units $\GELU$, $\rho=0.75$) and an even $50\%$-named split ($\rho=0.5$). Both place
\emph{every} attention value under Eq.~\ref{eq:boolean}. These are the models in
which both halves of the transformer---``what it moves'' and ``what it combines''---are named operations at
once.

\subsection{Training and evaluation}
All models are $125$M-parameter, $12$-layer, $768$-wide transformers with learned positional embeddings,
trained for one epoch on the same open-web corpus and tokenizer as the baseline, with identical optimizer
and schedule. The conventional \textsc{gelu} baseline is architecturally identical except for the
value transform and FFN. We evaluate language-model quality with LAMBADA \citep{paperno2016lambada}, BLiMP
\citep{warstadt2020blimp}, and ARC-Easy \citep{clark2018arc} via the standard harness
\citep{gao2023lmeval}---WinoGrande \citep{sakaguchi2020winogrande} sits at chance for all models and is
omitted---and we probe legibility directly on held-out text (defined in Section~\ref{sec:results-attn}). All language-model results are single-seed at this scale; we note this
where it matters.

\section{Specialized-Attention Designs: Legibility and Parity}
\label{sec:results-attn}

We first study the value-head designs in isolation, with a conventional feed-forward layer, so that any
change is attributable to the attention. We ask two questions: are the bounded values \emph{readable}, and
do they \emph{cost} anything? The answer to both turns out to depend on \emph{which} selectivity pressure is
applied (\S\ref{sec:selectivity}), and---this is the section's central finding---the right choice is
\emph{opposite} for the two designs: the single membership value wants both pressures, whereas the
conjunctive Boolean operator wants crispness alone.

\subsection{A legibility metric for value channels}
\label{sec:metric}
We probe each bounded value channel on held-out text ($8\times256$ tokens per batch, $12$ batches). We read
the head on two levels: the \emph{combined value} it feeds through $W_O$ (the membership $\sig(XW_v^h)$, or
the block $[A\cap B;A\setminus B]$ for a Boolean head), and the \emph{operands} it is built from (the
membership itself, or the two operand banks $A,B$ of Eq.~\ref{eq:boolean})---the atoms we actually want to
name.

For channel $c$ of special head $h$ we recompute its combined value at every token and record the mean
activation, the variance across tokens, and a \emph{crispness}---the fraction of tokens at which the value is
within $0.1$ of a rail ($\min(v,1-v)<0.1$), i.e.\ decisively on or off. We then classify each channel:
\begin{itemize}
\item \textbf{CONSTANT} (variance $<0.003$): the channel barely varies with context. This is the failure
mode of the vocabulary-tied design---a channel that has collapsed to a bias and detects nothing---and, as we
will see, also the mode into which an \emph{over-pressured} operator collapses.
\item \textbf{SELECTIVE} (variance $\geq0.003$ \emph{and} crispness $>0.5$): the channel varies with context
\emph{and} is decisively on/off---a crisp, contextually driven detector, the legible outcome.
\item \textbf{FUZZY} (the remainder): varies but is not crisp---a soft gate rather than a readable detector.
\end{itemize}
SELECTIVE is the fraction we care about: a channel that is both crisp and contextual is one we can name by
its max-activating contexts. Because it requires \emph{both} variance and crispness, it is a single number
that fails in two distinguishable ways---toward CONSTANT (crisp but dead) or toward FUZZY (contextual but
indecisive).

To separate those two failure modes we track two further statistics on the \emph{operands}, which the
three-regime sweep below makes central:
\begin{itemize}
\item \textbf{atom crispness}: the fraction of operand values (the membership, or $A$ and $B$ individually)
within $0.1$ of a rail. This measures whether the \emph{atoms}---not the combined output---are decisive, and
it is the quantity the crispness pressure targets directly.
\item \textbf{ON-fraction}: the fraction of operand values decisively \emph{on} ($v>0.9$). A low
ON-fraction with high crispness means the operands are crisply \emph{off} almost everywhere---the operator is
alive on paper but rarely fires. We read ON-fraction as ``how alive the operators are'': a healthy detector
is off by default but must actually turn on somewhere.
\end{itemize}

\subsection{Legibility is a trained feature}
\label{sec:legib-regimes}
Table~\ref{tab:attn-legib} reports the four statistics for every design under each selectivity regime. Read
by columns, three facts stand out, and together they are the section's central result.

The crispness term raises atom crispness
sharply and monotonically. For the Boolean operands it climbs from $32$--$33\%$ under sparsity alone to
$53$--$84\%$ once the crispness term is present (either alone or with sparsity); the membership operands,
already fairly crisp under sparsity, hold at $74$--$86\%$. On its own target metric the crispness term does
exactly what \S\ref{sec:selectivity} promises: it drives the atoms to the rails.

Crisp atoms are necessary but not sufficient: under \emph{both} pressures together the Boolean operators
over-sparsify into a constant-off collapse. The Boolean ON-fraction falls to
$0.2$--$0.9\%$: the operands are crisp, but crisp \emph{off}---they almost never fire. The mechanism is
specific to a conjunction. The value $A\cap B$ requires two operands high at once, so the intersection
already fires only rarely; add a sparsity term that pushes both operands \emph{down} toward $0$ and the
conjunction is starved to the point of never firing, at which point the channel stops varying and is read as
CONSTANT. The metric confirms the collapse: SELECTIVE drops from $60$--$62\%$ under sparsity alone to
$38$--$54\%$ under both pressures, while CONSTANT rises correspondingly (to $58\%$ at \bvh{}-50). It also
\emph{worsens over training}: for \bvh{}-100 under both pressures SELECTIVE falls from $60\%$ at $50$k steps
to $54\%$ by the end of the epoch, as the sparsity term keeps thinning the operands the longer it acts.

Removing the sparsity term averts the collapse, and crispness alone keeps the operators alive.
Under crispness alone the Boolean ON-fraction stays at $12$--$19\%$---an order of magnitude above the
both-pressures regime---and SELECTIVE recovers to $44$--$56\%$. The crispness term supplies decisiveness
without the downward pull that starves the conjunction, so the operators stay lively \emph{and} crisp. This
is the empirical form of the argument in \S\ref{sec:selectivity}: for a conjunctive operator the
intersection supplies its own sparsity, so an added sparsity pressure is not merely redundant but actively
harmful.

The membership head is the exception, because for a single value sparsity \emph{is} the selectivity: with no
conjunction to supply sparsity, the logic reverses. Here both pressures together are best:
\mvh{}-100 reaches its highest SELECTIVE ($59\%$) under both pressures, above sparsity alone ($44\%$) and
crispness alone ($46\%$), and---uniquely---crispness \emph{alone} is the worst of the two single pressures
for the combined-value metric. For a lone value the sparsity term is what carves ``fires on a specific set of
contexts'' out of ``fires softly everywhere''; there is no conjunction for it to starve, so it is pure
benefit. The same pressure that kills the Boolean operator makes the membership value.

\begin{table}[t]
\centering
\small
\begin{tabular}{llcccc}
\toprule
Design & pressure & atom-crisp & ON\% & SELECTIVE & CONSTANT \\
\midrule
\multicolumn{6}{l}{\emph{Membership head}}\\
\mvh{}-50  & sparsity alone  & 77\% & 0.1  & 58\% & 30\% \\
           & crispness alone & 84\% & 0.9  & 53\% & 41\% \\
           & both pressures  & 86\% & 0.1  & 57\% & 39\% \\
\mvh{}-100 & sparsity alone  & 74\% & 3.3  & 44\% & 35\% \\
           & crispness alone & 74\% & 5.3  & 46\% & 36\% \\
           & both pressures  & 81\% & 0.1  & \textbf{59\%} & 34\% \\
\addlinespace
\multicolumn{6}{l}{\emph{Boolean head}}\\
\bvh{}-50  & sparsity alone  & 32\% & 1.5  & 60\% & 32\% \\
           & crispness alone & 73\% & 19.4 & 44\% & 47\% \\
           & both pressures  & 84\% & 0.2  & 38\% & 58\% \\
\bvh{}-75  & sparsity alone  & 33\% & 1.8  & 62\% & 31\% \\
           & crispness alone & 67\% & 17.0 & 48\% & 42\% \\
           & both pressures  & 75\% & 0.4  & 45\% & 47\% \\
\bvh{}-100 & sparsity alone  & 32\% & 2.2  & 60\% & 27\% \\
           & crispness alone & 53\% & 12.1 & 56\% & 32\% \\
           & both pressures  & 63\% & 0.9  & 54\% & 37\% \\
\bottomrule
\end{tabular}
\caption{Value-channel legibility across the three selectivity regimes (Section~\ref{sec:metric}). \emph{atom-crisp}
and \emph{ON\%} are measured on the operands (the atoms we name); \emph{SELECTIVE}/\emph{CONSTANT} on the
combined value. Crispness pressure raises atom-crispness, but for the Boolean head \emph{both} pressures
over-sparsify the operators into a constant-off collapse (ON$\,\to\,0.2$--$0.9\%$, SELECTIVE $60\%\!\to\!38$--$54\%$),
which crispness \emph{alone} averts (ON $12$--$19\%$). For the membership head the reverse holds: both
pressures give the highest SELECTIVE.}
\label{tab:attn-legib}
\end{table}

\begin{figure}[t]
\centering
\begin{subfigure}{0.49\textwidth}
\includegraphics[width=\linewidth]{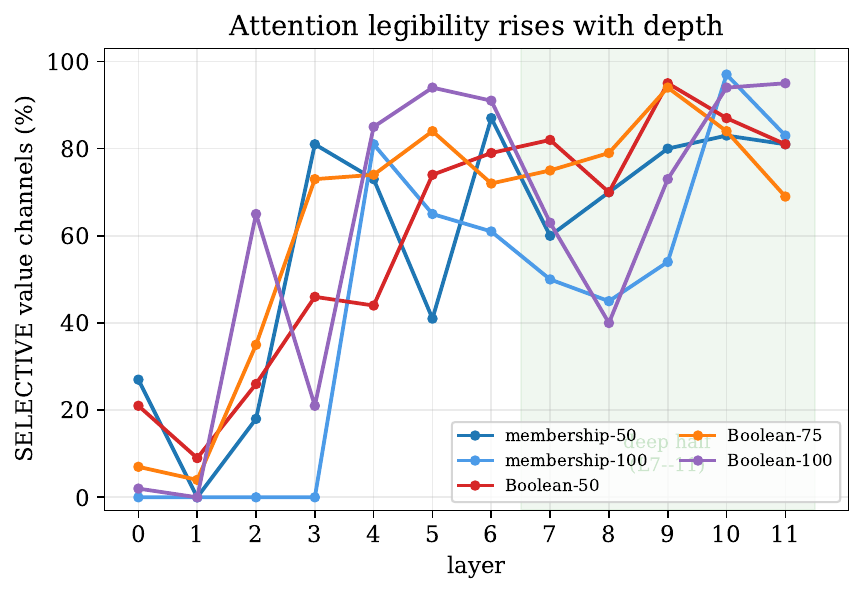}
\caption{SELECTIVE value channels by layer.}
\label{fig:attn-depth}
\end{subfigure}\hfill
\begin{subfigure}{0.49\textwidth}
\includegraphics[width=\linewidth]{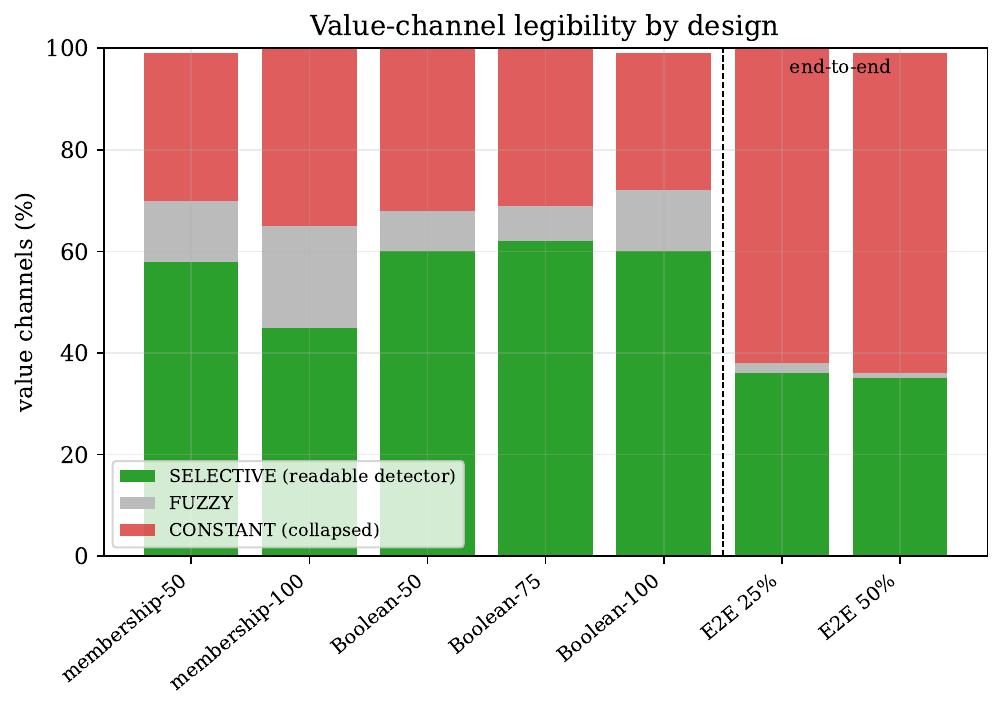}
\caption{Legibility composition by design.}
\label{fig:legib-summary}
\end{subfigure}
\caption{(a) Attention legibility rises with depth in every design; the deep half of the network reaches
$70$--$95\%$ selective. (b) Composition of value channels across the specialized-attention designs and the two
end-to-end models; SELECTIVE (green) dominates in the specialized-attention designs, while the end-to-end
models push attention legibility deeper and shallower attention toward CONSTANT (Section~\ref{sec:results-e2e}).}
\label{fig:legib}
\end{figure}

Legibility still rises with depth, orthogonal to the pressure choice: \emph{where} the legibility
lives is unchanged from design to design. Figure~\ref{fig:attn-depth} plots SELECTIVE\% by layer: in every
design the fraction climbs with depth, with shallow layers noisy or near-constant and the deep half (L7--11)
far more legible, reaching $70$--$95\%$ SELECTIVE. This contradicts the folklore that attention heads
crystallize only on low-level cues like punctuation and position: that folklore is a statement about
\emph{shallow} heads read in the \emph{vocabulary} basis, and when we read the deep value in its own bounded
basis it crystallizes readily. It also frames the collapse above as a story about \emph{how many} deep
detectors survive the pressure, not about \emph{where} they sit.

\subsection{The surviving detectors read as features, not punctuation}
Table~\ref{tab:detectors} lists representative SELECTIVE channels from the operator-alive regimes (sparsity
alone and crispness alone), sampled at a middle layer (L6) and named by their max-activating contexts. They
are recognizable: an abstract-concept channel (\emph{body, mind, love, idea}), a domain channel (\emph{Health,
military, generals, philosophy}), a people/plural channel (\emph{everyone, users, politicians}), a topical
channel (\emph{businesses, taxes, concerns}), alongside the expected clause-boundary punctuation detector.
As with the FFN units of \citet{oskin2026ncffn} and with sparse-autoencoder features, there is a
polysemantic tail---roughly half of the channels read cleanly---but the presence of crisp, contextual,
\emph{semantic} detectors already at a middle layer is exactly what the QK/OV folklore says should not be there.

The contrast with the collapsed regime is qualitative as well as quantitative. In the operator-alive regimes
the surviving SELECTIVE channels are nameable features like these. Under both pressures, where the Boolean
ON-fraction has fallen to near zero, the operators are dead and the few channels that still vary lock onto
high-frequency function words rather than content---consistent with a value that has been squeezed until only
the cheapest, most frequent distinctions survive. The named-detector yield and the ON-fraction move together:
keeping the operators alive is what keeps the channels nameable.

\begin{table}[t]
\centering
\small
\begin{tabular}{llll}
\toprule
Design & channel & max-activating tokens & reading \\
\midrule
\mvh{}-100 & L6 h4c48 & \emph{body, mind, love, point, idea} & abstract concept \\
\mvh{}-100 & L6 h10c4 & \emph{Health, military, generals, philosophy} & domain/topic \\
\mvh{}-50  & L6 h5c2  & \emph{businesses, taxes, concerns, business} & economy topic \\
\bvh{}-100 & L6 h3c31 & \emph{everyone, her, users, politicians} & people/plural \\
\bvh{}-100 & L6 h10c2 & \texttt{; ). \textbackslash n .} & clause boundary \\
\bvh{}-75  & L6 h1c30 & \emph{regardless, going, the, not} & (polysemantic tail) \\
\bottomrule
\end{tabular}
\caption{Representative SELECTIVE value channels (layer 6) from the operator-alive regimes, named by their
highest-activating held-out contexts. They read as semantic and grammatical features, not merely punctuation.
About half read cleanly; the rest are polysemantic, as in comparable sparse-feature analyses. When the
operators are instead starved into the constant-off collapse, this named-detector yield disappears.}
\label{tab:detectors}
\end{table}

\subsection{Language-model quality}
Table~\ref{tab:bench-attn} gives downstream numbers across all three regimes against the conventional
baseline. The headline is \emph{parity}: nearly every configuration sits in the $0.22$--$0.27$ LAMBADA band
around the baseline's $0.2536$, with BLiMP within $0.03$ and ARC-Easy within $0.05$, so legibility does not
come at a measurable aggregate quality cost. WinoGrande is at chance for every model and is omitted. Two real
movements survive that parity band, and they are movements in \emph{quality} produced by the
same pressure choices that drove the legibility result.

\begin{table}[t]
\centering
\small
\begin{tabular}{llcccc}
\toprule
Design & pressure & LAMBADA & L-ppl$\,\downarrow$ & BLiMP & ARC-E \\
\midrule
\textsc{gelu} baseline & --- & 0.2536 & 109.4 & 0.8066 & 0.4120 \\
\midrule
\multicolumn{6}{l}{\emph{Membership head}}\\
\mvh{}-50  & sparsity alone  & 0.2649 & 110.0 & 0.7982 & 0.3990 \\
           & crispness alone & 0.2257 & 149.1 & 0.7974 & 0.4036 \\
           & both pressures  & 0.2567 & 112.5 & 0.8070 & 0.3880 \\
\mvh{}-100 & sparsity alone  & \textbf{0.2142} & 189.6 & 0.7907 & 0.3830 \\
           & crispness alone & 0.2220 & 209.6 & 0.7767 & 0.3708 \\
           & both pressures  & \textbf{0.2313} & 146.2 & 0.8032 & 0.3788 \\
\addlinespace
\multicolumn{6}{l}{\emph{Boolean head}}\\
\bvh{}-50  & sparsity alone  & \textbf{0.2727} & 94.4  & 0.7982 & 0.3902 \\
           & crispness alone & 0.2538 & 116.2 & 0.7974 & 0.4061 \\
           & both pressures  & 0.2480 & 106.2 & 0.7944 & 0.4040 \\
\bvh{}-75  & sparsity alone  & 0.2383 & 123.0 & 0.7951 & 0.4066 \\
           & crispness alone & 0.2366 & 131.1 & 0.8017 & 0.4070 \\
           & both pressures  & 0.2527 & 118.4 & 0.7974 & 0.3939 \\
\bvh{}-100 & sparsity alone  & 0.2468 & 129.6 & 0.8050 & 0.3986 \\
           & crispness alone & 0.2368 & 144.4 & 0.8042 & 0.3851 \\
           & both pressures  & 0.2368 & 142.6 & 0.7995 & 0.3939 \\
\bottomrule
\end{tabular}
\caption{Downstream accuracy across the three selectivity regimes (single seed, one epoch, full evaluation
sets; higher is better except L-ppl, LAMBADA perplexity, lower is better). Bold marks the departures that
survive the parity band: \mvh{}-100 is a genuine loss under sparsity alone that \emph{recovers} under both
pressures, and \bvh{}-50 is a genuine LAMBADA gain.}
\label{tab:bench-attn}
\end{table}

The first movement is the \mvh{}-100 recovery, and it mirrors the legibility story exactly. When
\emph{every} head's value is squeezed to $[0,1]$, the bottleneck bites: under sparsity alone \mvh{}-100 drops
to LAMBADA $0.2142$ with perplexity $189.6$, the paper's clearest quality loss. Adding the crispness term
turns this into the \emph{best} membership configuration---LAMBADA $0.2313$, perplexity down to $146.2$, and
the highest membership BLiMP at $0.8032$---the same both-pressures regime that gave \mvh{}-100 its highest
SELECTIVE. For a single bounded value, the two pressures help both legibility and quality at once; there is no
tradeoff to manage. (At half strength, \mvh{}-50, the six unbounded heads retain enough headroom that the
sparsity-alone cost never appears.)

The second movement is \bvh{}-50 under sparsity alone (LAMBADA $0.2727$, $+0.019$ over baseline), and its
main lesson is negative: a $4$-layer pilot in which the \emph{fully}-Boolean design was best does not survive
scaling. At $12$ layers \bvh{}-100 regresses to parity (LAMBADA $0.2468$, BLiMP $0.8050$ versus baseline
$0.8066$) and the clearest gain moves to the \emph{mix}. Across the Boolean rows, quality is otherwise a
wash---no regime is reliably better than another on LAMBADA once the parity band is acknowledged---so unlike
the membership case the Boolean pressure choice is decided on \emph{legibility}, not quality. Both tables want
a seed sweep before any ordering within the parity band is trusted; a single seed does not support
an ordering among movements this small.

\subsection{Which pressure for which design}
The two tables agree on a single, actionable rule, and it is not the ``more pressure is better'' one might
expect. For a \emph{single membership value}, keep \emph{both} pressures: sparsity is what makes a lone value
selective, and here it improves legibility and quality together. For a \emph{conjunctive Boolean operator},
use \emph{crispness alone}: the intersection already supplies its own sparsity, so an added sparsity term does
not sharpen the operator but over-sparsifies it---starving both operands toward zero until the conjunction
never fires and the channel collapses to constant-off, a collapse that deepens the longer training runs. The
mechanism is the same object seen from two sides: the sparsity pressure removes coordinates, which is exactly
what a lone value needs and exactly what a conjunction cannot afford. This design-dependent verdict is what we
carry into the end-to-end models of Section~\ref{sec:results-e2e}.

\section{An End-to-End Legible Language Model}
\label{sec:results-e2e}

We now couple the Boolean attention (every value under Eq.~\ref{eq:boolean}, at $100\%$)---the more selective
of the two value designs (Section~\ref{sec:results-attn}), and the one that makes the value an explicit
logical operation---to the legible feed-forward layer of \citet{oskin2026ncffn}, whose units are fuzzy set
operations plus a self-forgetting sequence quantifier. This is the first model in which both halves of the
transformer---what the attention \emph{moves} and what the feed-forward layer \emph{combines}---are named
operations at once, and the question is whether the two legible components compose, in quality and in
legibility, or interfere.

Two design axes stay open, and we sweep both. The feed-forward layer exposes an \emph{operator fraction} (a
$\GELU$ partition, the $\rho$ of \citealp{oskin2026ncffn}): we run a $25\%$-named split and an even
$50\%$-named split. And, central to this section, each legible surface can carry a
\emph{selectivity pressure}
(Section~\ref{sec:selectivity}), and the pressure on the attention need not match the one on the feed-forward
layer. Every end-to-end variant therefore has an attention pressure regime and, optionally, a feed-forward
pressure regime; we describe a variant as placing selectivity \emph{on the attention only} (a plain
feed-forward layer beneath it) or \emph{on attention and the feed-forward layer}. The headline is that the
\emph{crispness-alone} variants---the pressure that makes operands decisive without also sparsifying them into
dead constants---keep the feed-forward operators \emph{alive} and namable as a standing bank, so that a
prediction can be read off named units (Figure~\ref{fig:combined}, Figure~\ref{fig:dla}); the quality-strongest
variant applies both pressures to the feed-forward layer and sits at the top of the parity band, but leaves most of its
operators crisp yet dead, few of them alive to name. The quality-versus-legibility tradeoff of the whole
design lives on the feed-forward surface, and we make it explicit below.

\subsection{Every variant trains}
The companion paper locates divergence just above an even Boolean split, and
Appendix~\ref{sec:trainability} finds an analogous boundary in the membership family, so the even-split
configuration---an even feed-forward split \emph{plus} fully-Boolean attention on top---was the run we expected
to diverge. It did not. Every end-to-end variant in this section trained a full epoch cleanly, across both
operator fractions and all three pressure regimes. Cumulative training perplexity is flat at $\approx\!29$--$30$
for all of them and carries no signal; the discriminating axes are downstream (\S\ref{sec:e2e-parity}). The
Boolean attention, at $100\%$, carries no starved-minority risk of its own (Appendix~\ref{sec:trainability})
and did not push the feed-forward layer over its horizon. The one coupling that \emph{does} diverge---a more
aggressive, fully-Boolean feed-forward variant under both pressures, where crispness deepens the operator's
saturation past its stall horizon---is a different and stricter feed-forward layer than the legible one studied
here; it is reported in Appendix~\ref{sec:trainability}, and its crispness-alone counterpart trains.

\subsection{Parity holds across the end-to-end sweep}
\label{sec:e2e-parity}
Table~\ref{tab:bench-e2e} gives downstream quality. Because run perplexity is flat, LAMBADA and BLiMP are the
discriminating axes, and every end-to-end variant sits in or above the parity band on both: the two most
aggressive legibility constraints we have---bounded Boolean attention values and named feed-forward
operators---compose without breaking quality. The strongest variant is the even-split model with both pressures
on attention \emph{and} the feed-forward layer, at LAMBADA $0.2657$ and BLiMP $0.8163$ against the baseline's
$0.2536$ and $0.8066$; with a single seed and margins this small we read it as sitting at the top of the parity
band, not as a reliable win. The uniform slip on ARC-Easy ($\leq0.04$) is the same
parity-band movement seen for the specialized-attention designs---single-seed
texture, not a reliable ordering.

\begin{table}[t]
\centering
\small
\begin{tabular}{llccc}
\toprule
Selectivity / regime & Frac. & LAMBADA & BLiMP & ARC-E \\
\midrule
\textsc{gelu} baseline & --- & 0.2536 & 0.8066 & 0.4120 \\
\midrule
\multicolumn{5}{l}{\emph{Selectivity on the attention only (plain feed-forward)}} \\
sparsity alone   & $25\%$ & 0.2544 & 0.8039 & 0.3977 \\
sparsity alone   & $50\%$ & 0.2449 & 0.8027 & 0.3923 \\
both pressures   & $25\%$ & 0.2575 & 0.8061 & 0.3918 \\
both pressures   & $50\%$ & 0.2488 & 0.8004 & 0.3977 \\
crispness alone  & $25\%$ & 0.2511 & 0.8103 & 0.3939 \\
crispness alone  & $50\%$ & 0.2451 & 0.8104 & 0.3931 \\
\midrule
\multicolumn{5}{l}{\emph{Selectivity on attention and the feed-forward layer}} \\
sparsity alone   & $25\%$ & 0.2544 & 0.8095 & 0.3893 \\
sparsity alone   & $50\%$ & 0.2519 & 0.8047 & 0.3902 \\
both pressures   & $25\%$ & 0.2608 & 0.8103 & 0.3737 \\
both pressures   & $50\%$ & \textbf{0.2657} & \textbf{0.8163} & 0.3939 \\
crispness alone  & $25\%$ & 0.2472 & 0.8062 & 0.3939 \\
crispness alone  & $50\%$ & 0.2612 & 0.7978 & 0.3986 \\
\bottomrule
\end{tabular}
\caption{End-to-end quality (accuracy; higher is better). Every variant places \emph{every} attention value
under Eq.~\ref{eq:boolean} at $100\%$ and names a fraction of feed-forward units; ``Frac.'' is the named
operator fraction. Run perplexity is flat ($\approx\!29$--$30$) across all variants and is omitted; WinoGrande
is at chance for all models and is omitted. The even-split both-pressures model (bold) is nominally highest on
LAMBADA and BLiMP, at the top of the parity band; with a single seed the small margin over baseline is not a
reliable win. Single-seed, one epoch.}
\label{tab:bench-e2e}
\end{table}

\subsection{Legibility, surface one: the feed-forward operands}
\label{sec:e2e-ffn-legib}
The parity numbers say the components coexist; legibility says what one can read off them. We start with the
feed-forward layer, because this is where the end-to-end legibility turns out to live, and it is the central
new result of this section. Table~\ref{tab:ffn-legib} measures the Boolean operands $A,B$ of the feed-forward
operators with the same statistics used for the attention value (Section~\ref{sec:results-attn}): the fraction
of operand tokens within $0.1$ of a rail (crispness), the fraction of operands above $0.9$ (``ON'', i.e.\ the
operator is alive rather than saturated-off), and the SELECTIVE fraction (crisp \emph{and} contextual
channels).

Three things follow, and they mirror the attention story exactly. First, a \emph{plain} legible feed-forward
layer has barely legible operands---$\approx\!50\%$ crisp and $\approx\!0\%$ SELECTIVE---which is the
companion paper's ``polysemantic operands'' limitation, now quantified. Second, a sparsity pressure applied to
the \emph{combined} operator output is inert for the atoms: it moves operand crispness only $50\%\!\to\!51\%$
at $0\%$ SELECTIVE, because penalizing $A\cap B$ does nothing to $A$ or $B$---you must target the operands
themselves. Third, an operand-targeted pressure crispens them dramatically ($50\%\!\to\!81$--$95\%$), and here
the same split as the attention returns: both pressures leave the operands crisp but nearly \emph{dead}
(ON $1.9$--$3.5\%$), while crispness \emph{alone} leaves them crisp and \emph{alive} (ON $17$--$33\%$) and gives
the best SELECTIVE fraction ($25\%$ at the $25\%$ operator setting). Crispness alone is what makes the
feed-forward operators readable.

\begin{table}[t]
\centering
\small
\begin{tabular}{llcccc}
\toprule
Feed-forward pressure & Frac. & op-crisp & ON\% & SELECTIVE & CONSTANT \\
\midrule
none (plain)                     & $25\%$ & $50$ & $0.9$  & $1$          & $58$ \\
sparsity (on operator output)    & $25\%$ & $51$ & $0.8$  & $1$          & $50$ \\
both pressures (on operands)     & $25\%$ & $95$ & $3.5$  & $13$         & $85$ \\
crispness alone (on operands)    & $25\%$ & $91$ & $32.7$ & $\mathbf{25}$ & $70$ \\
\midrule
none (plain)                     & $50\%$ & $50$ & $0.6$  & $0$          & $52$ \\
sparsity (on operator output)    & $50\%$ & $51$ & $0.5$  & $0$          & $55$ \\
both pressures (on operands)     & $50\%$ & $86$ & $1.9$  & $15$         & $73$ \\
crispness alone (on operands)    & $50\%$ & $81$ & $17.5$ & $11$         & $69$ \\
\bottomrule
\end{tabular}
\caption{Legibility of the feed-forward operands (percentages; ``Frac.'' is the named operator fraction).
op-crisp is the fraction of operand tokens within $0.1$ of a rail; ON the fraction above $0.9$ (operators
alive); SELECTIVE the crisp-and-contextual fraction; CONSTANT the crisp-but-context-free fraction. A plain
layer is illegible ($\approx\!50\%$ crisp, $\approx\!0\%$ SELECTIVE); sparsity on the \emph{combined} output is
inert; only operand-targeted pressure crispens the atoms, and crispness alone keeps them alive and is the most
SELECTIVE (bold).}
\label{tab:ffn-legib}
\end{table}

Because crispness alone keeps the operators alive, its living units can be named. Table~\ref{tab:ffn-units}
lists the most-engaged operators of a crispness-alone model, named by their max-activating held-out contexts:
sentence-initial discourse markers and conditional openers early, an infinitival/modal-complement detector in
the middle, and---in the final layer---clean \emph{comparative}, \emph{negation}, and \emph{perfect-aspect}
units. This is the companion paper's ``named logical operator'' outcome reproduced with legible attention
stacked on top, and it is exactly the both-pressures variant \emph{cannot} give: crisp operators that never
fire are crisp on paper but carry nothing to name.

\begin{table}[t]
\centering
\small
\begin{tabular}{llll}
\toprule
FFN unit & crispness & top-activating tokens & reading \\
\midrule
L1 u18  & 1.00 & \emph{there, Surely, Indeed, Meanwhile, No} & sentence-initial/discourse \\
L1 u57  & 0.99 & \emph{( , if, as, whether, when, If, which} & conditional/subordinator \\
L6 u49  & 1.00 & \emph{to, think, need, insists} & infinitival/modal complement \\
L11 u39 & 0.98 & \emph{less, more, faster, -ier, different} & \textbf{comparative} \\
L11 u89 & 0.98 & \emph{not, nothing, even, non} & \textbf{negation} \\
L11 u11 & 0.95 & \emph{had, has, have, having, since} & \textbf{perfect aspect} \\
\bottomrule
\end{tabular}
\caption{Most-engaged feed-forward units of a crispness-alone end-to-end model, named by max-activating
held-out contexts. The final-layer triple---comparative, negation, aspect---is the ``named logical operator''
outcome, now inside a model whose attention is also constrained.}
\label{tab:ffn-units}
\end{table}

\subsection{Legibility, surface two: the attention, and a division of labor}
\begin{figure}[t]
\centering
\includegraphics[width=0.62\textwidth]{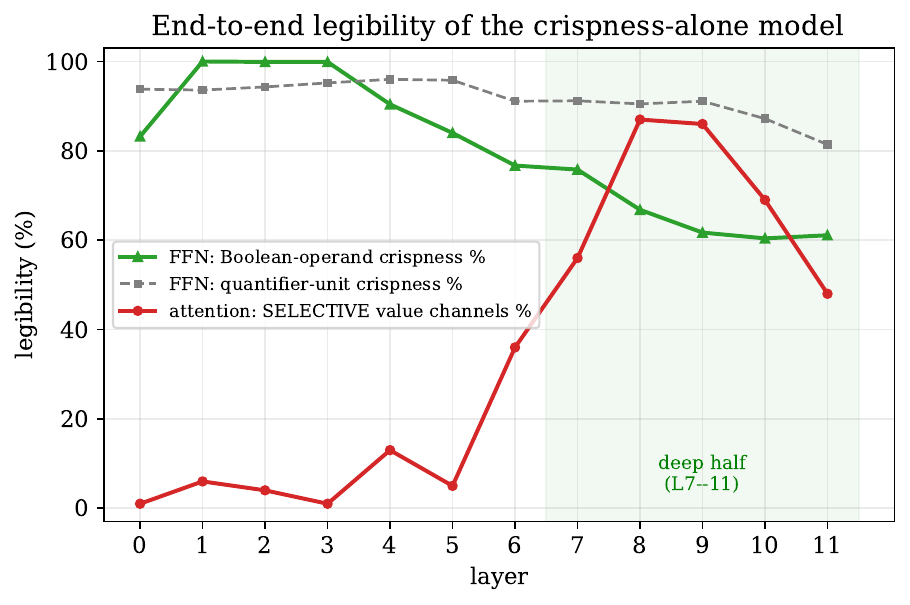}
\caption{End-to-end legibility by layer, for a crispness-alone model. The feed-forward operands are crisp at
every depth (\S\ref{sec:e2e-ffn-legib}), while the Boolean attention value is selective only in the deep half
and goes constant in the shallow layers---the division of labor discussed in the text: the legible
feed-forward layer takes the shallow work, so the attention need only be readable at depth.}
\label{fig:combined}
\end{figure}

Figure~\ref{fig:combined} reads both surfaces of an end-to-end model layer by layer. The feed-forward side is
legible throughout, as just shown; the attention side tells the expected caveat, and it is precisely the one
this paper predicts. Coupled to a legible feed-forward layer, the shallow attention goes \emph{constant}:
across both operator fractions the end-to-end attention is only $30$--$35\%$ SELECTIVE with $64$--$69\%$ of
channels CONSTANT, well below the $60\%$ SELECTIVE of standalone \bvh{}-100 (Figure~\ref{fig:legib-summary}).
Its operands are crisp ($63$--$73\%$), and, exactly as on the feed-forward surface, crispness alone keeps them
alive (ON $12\%$) where both pressures kill them (ON $0.3\%$). The interpretation is a division of labor: the
legible feed-forward layer absorbs the shallow, local work, freeing the attention to specialize---and be
readable---only at depth, where it must. The claim is therefore precise: the model is legible in its
feed-forward operands throughout and in its attention value at depth, not everywhere and not in shallow
attention. The end-to-end legibility lives in the feed-forward layer; the attention contributes a legible deep
fraction and a constant shallow one, and closing that gap is stated as future work in
Section~\ref{sec:discussion}.

\begin{figure}[t]
\centering
\includegraphics[width=\textwidth]{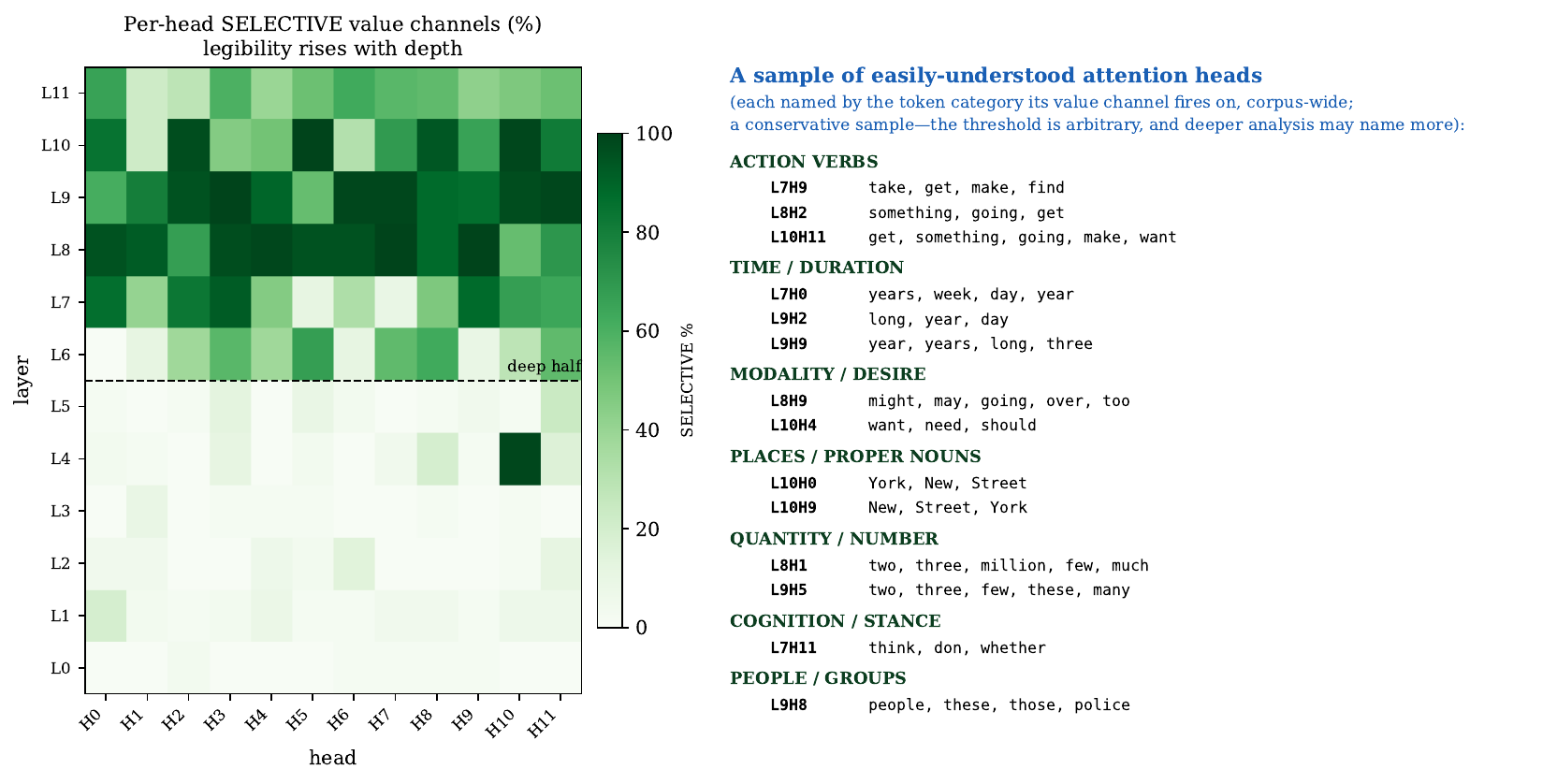}
\caption{Per-head attention legibility across all $144$ heads, measured over held-out corpus text---a
description of \emph{what each head detects across all inputs}, not of any single prediction. \emph{Left:} the
SELECTIVE fraction (crisp \emph{and} contextual value channels) of each head---near-zero in the shallow layers,
rising to $87$--$88\%$ by layers~$8$--$9$; the depth gradient of Figure~\ref{fig:combined}, resolved per head.
\emph{Right:} a \emph{sample} of the easily-understood deep heads---those whose most content-rich value channel
fires on a clearly nameable \emph{category} of tokens, grouped by category (modality, time, place names,
quantity, action verbs, cognition, people). We do not treat the size of this sample as a hard boundary: it
depends on an arbitrary readability threshold, and many of the remaining deep heads---crisp and contextual, but
reading as syntactic or functional patterns here---likely carry a nameable meaning too, only harder to surface,
which the deeper per-channel analysis we leave to future work may recover. On the sample itself, two caveats:
this is each head's \emph{most content-rich} channel (its others are often functional), and a naive
most-active-token readout is frequency-confounded (it collapses to ``the''/``people''), so the labels restrict
to the tokens a channel fires on decisively.}
\label{fig:head-legib}
\end{figure}

A per-head view sharpens the claim. Figure~\ref{fig:head-legib} (left) gives the SELECTIVE fraction of every
one of the $144$ heads: it is near-zero across the shallow layers and rises sharply through the deep half,
peaking at $87$--$88\%$ at layers~$8$--$9$ before easing at layer~$11$---the aggregate depth gradient, resolved
head by head. This heatmap has no baseline counterpart, and deliberately so: SELECTIVE is defined on a
\emph{bounded} value---crisp meaning within $0.1$ of a rail, contextual meaning it varies across tokens---so it
is simply undefined for a conventional head, whose value is unbounded and has no rails to be crisp against.
There is thus no baseline version of this figure to place beside it; the only legibility a standard model
exposes is the post-hoc, promote-side view of Figure~\ref{fig:dla-baseline}, and the detection-side selectivity
measured here exists only because the value is bounded by construction.
Reading a single prediction (Figure~\ref{fig:dla}) \emph{understates} this, because direct logit
attribution surfaces only the heads relevant to \emph{that} token; the legibility is a property of the standing
head bank, not of any one prediction. Naming \emph{what} a deep head detects is harder, and succeeds only in
part. A naive readout of a channel's most-active tokens collapses to frequent function words unless
one restricts to the tokens it \emph{decisively} fires on; under that stricter reading most SELECTIVE deep
heads resolve to syntactic or functional patterns (determiners, pronouns, particles), while a readily-nameable
sample of the deep heads carry a channel with a clearly nameable \emph{content} category
(Figure~\ref{fig:head-legib}, right)---place names (\emph{York}/\emph{New}/\emph{Street}), temporal spans
(\emph{year}/\emph{day}/\emph{week}), quantities (\emph{two}/\emph{million}/\emph{percent}), modality
(\emph{might}/\emph{want}/\emph{should}), action verbs (\emph{take}/\emph{get}/\emph{make}). The claim is
therefore graded: crisp, contextual legibility is near-universal in the deep half, and a cleanly nameable
content category is present in at least a readily-identifiable sample---the exact fraction is threshold-dependent
and a deeper per-channel analysis will likely name more, which we leave to future work---while, as on the
feed-forward surface, naming the operand's meaning remains the harder frontier.

\subsection{Legibility in action: reading a single prediction}
\begin{figure}[t]
\centering
\includegraphics[width=0.98\textwidth]{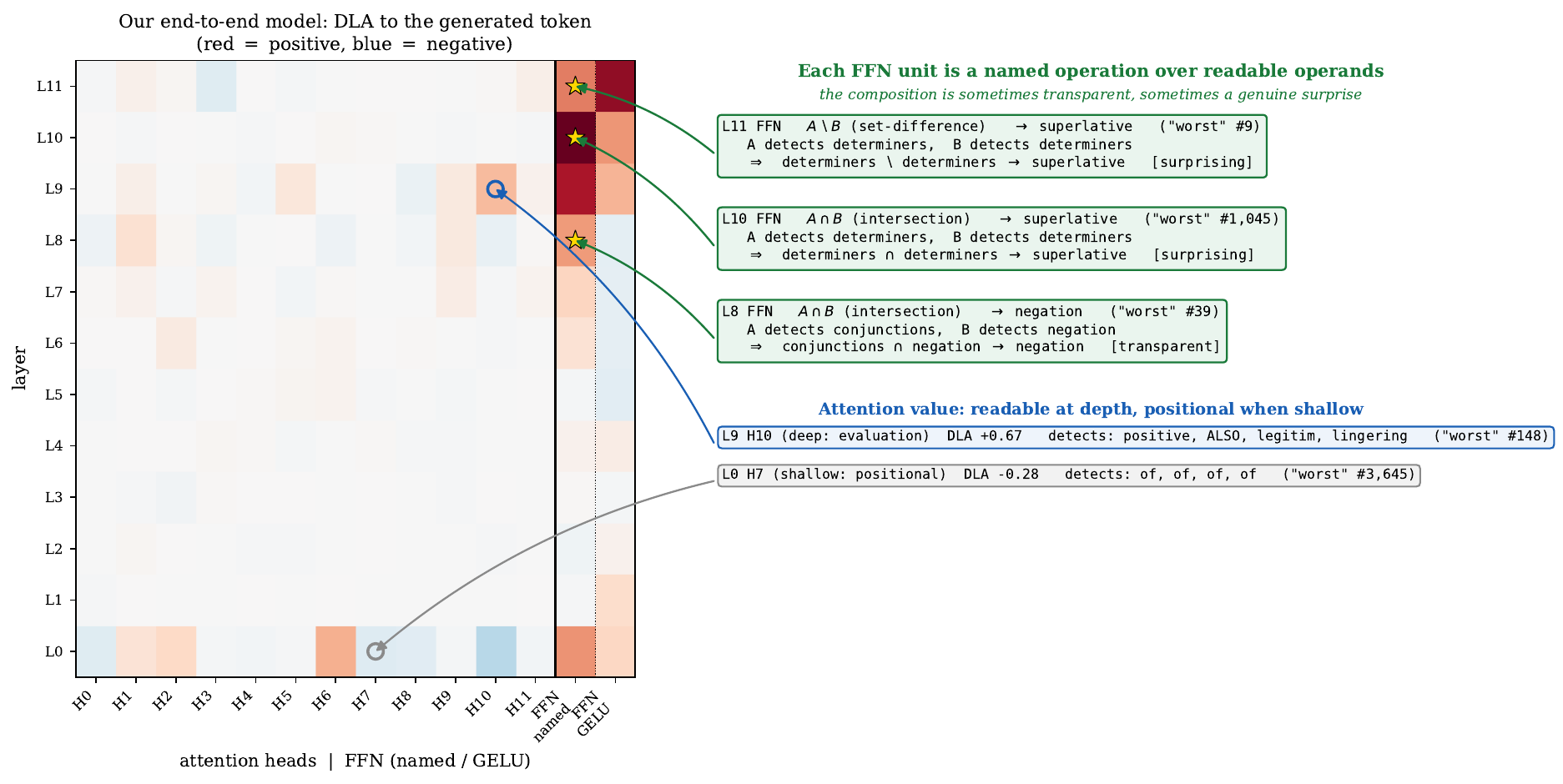}
\caption{Reading a single prediction on the crispness-alone headline model. \emph{Left:} direct logit attribution of the generated token
\emph{``worst''} (from \emph{``It was the best of times, it was the''}) across all $12$ layers, per attention
head and per feed-forward unit (red positive, blue negative). \emph{Right, top:} unlike the baseline's
monolithic block, each named feed-forward unit exposes not just \emph{what} it promotes but the \emph{operation
over which operands} produces it, and the operands read as functional features. Two of the three are a genuine
surprise---a superlative is a set-difference (or intersection) of two \emph{determiner} detectors,
$\mathrm{determiner}\setminus\mathrm{determiner}\to\mathrm{superlative}$ (layers~$11$ and $10$, \emph{worst} at
ranks $9$ and $1045$)---and one is transparent: a negation is an intersection one of whose operands is itself a
negation detector, $\mathrm{conjunction}\cap\mathrm{negation}\to\mathrm{negation}$ (layer~$8$, rank $39$). The
named form is what lets one see the composition at all. \emph{Right, bottom:} the top attention heads; the
attention contributes weakly (its readable deep head reaches $+0.67$ against the feed-forward's larger writes)
and splits by depth---the deep head reads the prompt's evaluative sentiment, the shallow head keys on position.
Three-quarters of the attribution is legible.}
\label{fig:dla}
\end{figure}

The aggregate statistics say the operators are readable; the payoff is reading them \emph{for a specific
prediction}, on the crispness-alone headline model. It completes \emph{``It was the best of times, it was
the''} with \emph{``worst''} (probability $0.83$). We decompose that token's logit by direct logit
attribution---writing it, through the tied unembedding, as an additive sum over every attention head and every
feed-forward unit---and name each contributor not by what it \emph{detects} but by the vocabulary
\emph{category} its write most promotes, together with the rank \emph{worst} holds in that promotion, because
the write, not the detected context, is the half that explains the prediction (Figure~\ref{fig:dla}).

Three-quarters of the token's attribution is \emph{legible}---carried by attention heads and named
feed-forward units rather than the opaque residual ($\GELU$) path---and its largest contributors are named
feed-forward units that compose. A \textsc{superlative} unit at layer~$11$ promotes
\emph{most}/\emph{greatest}/\emph{biggest}/\emph{least}/\emph{best}/\emph{highest}, placing \emph{worst} at
rank $9$ of $50{,}257$; a \textsc{negation} unit at layer~$8$ promotes \emph{not}/\emph{instead}/%
\emph{less} and their kin, placing \emph{worst} at rank $39$. The prediction reads directly off named units:
superlative axis $\cap$ negation $=$ \emph{worst}. The reading is not free of noise---of the six
largest contributors two carry \emph{worst} only weakly (rank $>10^3$) and the attention heads add little
here---but the nameable units dominate the attribution and suffice to explain the token. This is the concrete
form of what by-construction legibility buys: for an arbitrary token one can point at named writes and watch
them compose, without training an external probe. A methodological aside sharpens the thesis: a unit's
\emph{detect} signature (the contexts it fires on) is polysemantic, whereas its \emph{promote} signature (the
category it writes toward) is clean---a named \emph{operation} whose operands are fuzzy but whose \emph{effect}
is readable, and it is this promote-side legibility, preserved only while the operators stay alive, that the
aggregate tables track.

This is the sense in which the feed-forward layer goes \emph{beyond} the conventional one of
Figure~\ref{fig:dla-baseline}. There, the same superlative and negative-valence concepts are present, but each
is a single monolithic $\GELU$ block whose vocabulary promotion is recovered only \emph{after} training by a
lens. Here each contributor carries its logical \emph{form} on its face (Figure~\ref{fig:dla}): the superlative
and valence units are explicit set-differences---negations---and intersections, so one reads not only the
concept but the operation that composes it. And the feed-forward layer contains \emph{sequence-quantifier}
units, soft existential and proportional scans over the prefix, an operation a pointwise $\GELU$ network cannot
express at all. The baseline promotes a concept; ours exposes the operation that produces it.

The attention side of the same figure is worth reading, because for this token it says exactly what the
division of labor predicts. No head's attribution to \emph{worst} exceeds $|{+}0.77|$, against the
feed-forward units' $+2.14$: the attention barely moves this prediction, and the heads that do fire split
cleanly by depth. The shallow contributors are the textbook positional and punctuation heads---layer-$0$ heads
that key on \emph{of}, \emph{to}, or a newline and write toward commas and function words, the very ``heads
crystallize on punctuation'' pattern, here doing the local bookkeeping the feed-forward layer is free to
ignore. The readable \emph{deep} head, by contrast, is an \textsc{evaluation} detector that has read the
sentiment of the prompt: a layer-$9$ head fires on \emph{perfect}/\emph{excel}/\emph{love}/\emph{review}/%
\emph{well} and writes toward a valence axis (\emph{approval}/\emph{positive}/\emph{rejection}), having
registered ``the best of times'' as an evaluative context without itself promoting \emph{worst}. A single
prediction thus shows all three roles at once: the feed-forward units compose the token, the deep attention
supplies readable context, and the shallow attention supplies position.

\subsection{What the end-to-end model delivers}
The readable end-to-end model is the crispness-alone one. Its feed-forward operators are crisp \emph{and} alive
(Table~\ref{tab:ffn-legib}), its living units are namable (Table~\ref{tab:ffn-units}), and an arbitrary
prediction can be attributed to a composition of named operators without a post-hoc tool
(Figure~\ref{fig:dla})---all at quality inside the parity band. Its cost is modest and precisely located: the
both-pressures variant is the quality strongest (even split, LAMBADA $0.2657$ / BLiMP $0.8163$, at the top of
the parity band) but its feed-forward operators, crisp yet nearly dead (Table~\ref{tab:ffn-legib}), leave few units
\emph{alive} to name as a standing bank. That is the quality-versus-legibility tradeoff of the whole design,
and it lives on the feed-forward surface. What we
deliver is a language model at baseline quality whose dominant feed-forward computations are named operations
readable in the act of predicting a token, with legible attention at depth; what we do not yet deliver is a
network with no opaque part anywhere---the gap is the shallow attention (Section~\ref{sec:discussion}).
The reading at a second epoch, at matched training, is the same: benchmark parity. Continued for one more epoch,
with no change to the architecture or the selectivity pressures, the crispness-alone end-to-end models reach
LAMBADA $0.2635$ (perplexity $98.2$) and BLiMP $0.8086$ at the $25\%$-named split, and LAMBADA $0.2620$
(perplexity $94.6$) and BLiMP $0.7932$ at the even split (ARC-Easy $0.39$); the same \textsc{gelu} baseline
trained a second epoch reaches LAMBADA $0.275$ (perplexity $78.1$) and BLiMP $0.823$ \citep{oskin2026ncffn}. On
the discriminating axes---LAMBADA accuracy and BLiMP---the legible model and the baseline sit within the parity
band at both epochs (Table~\ref{tab:bench-e2e-ep2}); LAMBADA perplexity, the most seed-sensitive of the three,
moves more, and at two epochs the baseline is nominally ahead on all three aggregates by small margins. With a
single seed we cannot separate a real ordering from training-run and sampling noise, so we claim neither a gain
nor a loss at convergence: the reading is parity, and any drift toward the baseline at the second epoch
is as plausibly a single-seed artifact as a real effect. What the second epoch does \emph{not} do is disturb the
legibility, which is read off the same crispness-alone model throughout.

\begin{table}[t]
\centering
\small
\begin{tabular}{lcccccc}
\toprule
 & \multicolumn{3}{c}{Epoch 1} & \multicolumn{3}{c}{Epoch 2} \\
\cmidrule(lr){2-4}\cmidrule(lr){5-7}
Model & LAMBADA & L-ppl$\,\downarrow$ & BLiMP & LAMBADA & L-ppl$\,\downarrow$ & BLiMP \\
\midrule
\textsc{gelu} baseline \citep{oskin2026ncffn} & 0.2536 & 109.4 & 0.8066 & 0.2750 & 78.1 & 0.8230 \\
\midrule
crispness alone, $25\%$-named & 0.2472 & 115.7 & 0.8062 & 0.2635 & 98.2 & 0.8086 \\
crispness alone, even split   & 0.2612 & 109.9 & 0.7978 & 0.2620 & 94.6 & 0.7932 \\
\bottomrule
\end{tabular}
\caption{End-to-end quality at \emph{both} training epochs, against a \emph{matched} \textsc{gelu} baseline
(same architecture, corpus, and schedule; baseline numbers from \citealp{oskin2026ncffn}). On the discriminating
axes---LAMBADA accuracy and BLiMP---the legible model and the baseline sit within the parity band at both
epochs; at two epochs the baseline is nominally ahead by small margins on all three aggregates. With a single
seed we read this as benchmark parity rather than a reliable win or loss. Single seed, one scale. (The full
one-epoch sweep over pressures and operator fractions is Table~\ref{tab:bench-e2e}, where the strongest variant
sits at the top of the parity band on LAMBADA and BLiMP.)}
\label{tab:bench-e2e-ep2}
\end{table}

\section{Discussion}
\label{sec:discussion}

\paragraph{What the pieces say together.}
Read as one result, the paper makes a single architectural claim: a transformer can be built so that its
dominant computations are named operations, readable in their own basis, at no quality cost we can measure.
The feed-forward half was already legible \citep{oskin2026ncffn}; here the attention value becomes legible
too, under the same discipline---constrain what a component \emph{detects} and leave what it \emph{writes}
free---and the two compose into an end-to-end model at baseline perplexity and downstream accuracy. On the
attention side the constraint can be almost nothing---in its membership form, a single sigmoid on an otherwise
standard head---which makes the size of the payoff disproportionate to the change. The one
load-bearing lesson, learned the hard way from a design that collapsed, is that the site of the constraint is
everything: bound the pre-projection value and the head becomes a bank of selective detectors; pin the
post-projection write to the vocabulary and it collapses to the unigram prior. Legibility is not something
you add to a write; it is something you impose on a representation and then let the model use freely.
A second, quieter lesson comes from the operators that died when over-constrained: turning a bounded value into
a \emph{selective} one is not one knob but two---a sparsity pressure and a crispness pressure---and the right
mix is dictated by the operator rather than set globally. A lone membership value wants both; a conjunction,
already sparse because it fires only when both operands do, wants crispness alone and is starved to dead
constants if pushed to be sparser still. The pressures also have to land on the \emph{operands} one hopes to
name, not on the operator's combined output, where they do nothing.

\paragraph{Why by-construction legibility is worth the trouble.}
Post-hoc interpretability---lenses, probes, sparse autoencoders---has taught us most of what we know about
these models, and it will remain essential. But it has three structural costs that by-construction legibility
does not. It is \emph{external}: the explanation lives in a tool that must be trained and validated
separately, and can be wrong about the model. It is \emph{partial and contested}: which SAE, which width,
which layer, which threshold. And it is \emph{perishable}: every change to the model invalidates the tool and
demands it be rebuilt. A model whose units are named operations by construction pays none of these. The
comparative and negation units of Table~\ref{tab:ffn-units} are not hypotheses recovered by a dictionary that
might be mistuned; they are what those units \emph{are}. This does not make the model fully transparent---the
operands are still often polysemantic, and shallow attention is still opaque---but it moves the boundary of
what must be reconstructed inward, and it makes the reconstructed part smaller and better-posed (name the
\emph{operands} of a known operator, rather than discover both operator and operands at once).

\paragraph{What one can do with a legible-by-construction LLM.}
The point of reading a model is usually to \emph{act} on it, and named operations are directly actionable in
ways a dense activation is not.
\begin{itemize}
\item \textbf{Edit and steer by operator.} A membership channel or a Boolean/quantifier unit has a meaning
one can set, clamp, or ablate. Suppressing a negation unit, forcing a licensing detector on, or reading off
\emph{when} a comparative unit fires are single-unit interventions with a stated semantics---closer to editing
a program than to the diffuse activation-steering vectors current methods must search for.
\item \textbf{Audit and monitor.} Because the units are named, one can watch them at inference: raise a flag
when a particular detector fires, log which quantifiers are active on a given input, or gate on the presence
of a semantic feature. Safety-relevant monitoring becomes a matter of reading declared operators rather than
of training a probe to guess at hidden ones.
\item \textbf{Debug linguistic behavior.} The FFN's grammatical-licensing units and the attention's deep
semantic detectors give a direct handle on \emph{why} a model does or does not respect an agreement,
negation, or discourse constraint---a mechanism to inspect rather than a black box to prod.
\item \textbf{Cheaper, more faithful interpretation.} Half the work of a sparse autoencoder---discovering that
a unit computes an intersection or an existential---is done exactly and by construction, leaving only the
operands to name. Interpretation starts from a correct operator instead of from scratch.
\item \textbf{A substrate for verification.} Bounded memberships and explicit set operations are the kind of
object over which one can hope to state and check properties (monotonicity of a detector, an invariant on a
quantifier's reach). We do not attempt formal guarantees here, but a network of named fuzzy operations is a
far more promising target for them than a stack of $\GELU$ MLPs.
\end{itemize}
These are directions, not finished results; the contribution of this paper is the \emph{substrate} that makes
them approachable, together with evidence that adopting it costs nothing measurable in quality.

\paragraph{What the named operations reveal.}
A closer look at the operators' operands turns an apparent limitation into a finding. The two inputs $A,B$ of a
named feed-forward unit are not monosemantic, but neither are they noise: read with frequency correction they
resolve into recognizable \emph{functional} features---determiners and possessives, negation, modals,
conjunctions, copulas. What the named form then exposes, and a post-hoc lens on a monolithic block cannot, is
the \emph{operation the model runs over them}. Sometimes that composition is transparent: a unit that promotes
negation is an \emph{intersection} one of whose operands is itself a negation detector. Sometimes it is
genuinely non-obvious---the unit that promotes superlatives is a \emph{set-difference} of two \emph{determiner}
detectors, $\mathrm{determiner}\setminus\mathrm{determiner}\to\mathrm{superlative}$, a computation no one would
write down and that a conventional $\GELU$ block hides completely. We do not claim to understand why such a
composition works; the point is that a by-construction architecture lets us \emph{see} it at all---a different
and stronger kind of legibility than recovering the \emph{concept} a layer promotes
(Figure~\ref{fig:dla-baseline}).

\paragraph{Limitations.}
We are single-seed at one scale ($125$M, one epoch); the parity claim is robust but the within-band ordering
and the exact legibility percentages want a seed sweep, and scaling behavior is untested. Probing harder
prompts, moreover, suggests that some of what we see---the attention staying shallow, the crispest readings
falling on confident, feed-forward-driven tokens---is partly the ceiling of a small, shallow model; whether the
division of labor and the operator legibility survive at $1$B$+$ parameters is an open question we cannot
settle here. The attention
legibility is a \emph{deep-layer} phenomenon and, once coupled to a legible FFN, shallow attention goes
constant---so the end-to-end model is legible in its FFN throughout and its attention at depth, not
everywhere. The Boolean value is deliberately \emph{within-token}; genuine $n$-ary relational binding across
positions is a different mechanism we do not provide.

\paragraph{Future work.}
Four steps follow directly. \emph{Close the shallow gap}: understand why a legible FFN drives shallow
attention constant, and whether a shallow-specific legibility constraint (or simply accepting an opaque
shallow stem) is the right resolution. \emph{Remove the trainability boundary}: Appendix~\ref{sec:trainability}
points the cause at a saturated-product gradient, and a remedy aimed there---rather than at the residual
write---would allow a fully-Boolean FFN and a wider range of stable mixtures. \emph{Add relational binding} as
a separate, equally-legible mechanism, so that ``what relates to what'' joins ``what holds'' and ``what
combines.'' And \emph{demonstrate the actionability} above concretely---edit a negation unit and measure the
behavioral effect, monitor a detector under distribution shift---turning the substrate into evidence that
by-construction legibility pays off in use.

\section{Conclusion}
\label{sec:conclusion}
We set out to make attention legible by construction and to join it to an already-legible feed-forward layer.
The mechanism is one change---bound a head's value into fuzzy memberships (in its simplest form a single
sigmoid on an otherwise unchanged head), or into an explicit intersection and set-difference, and leave the
write free---and it turns the opaque half of the attention circuit into a bank of crisp, contextual detectors
whose legibility, contrary to folklore, \emph{grows} with depth. The same
choice that succeeds here is the mirror image of one that failed: pinning a head's output to the vocabulary
collapses it, while bounding what it detects and freeing what it writes does not. Turning a bounded value into
a \emph{selective} one takes two pressures---one toward firing sparsely, one toward firing crisply---and which
combination a design wants is itself part of the design: a single membership value wants both, whereas an
explicit conjunction, already sparse by construction, wants crispness alone and dies if pressed to be sparser.
Putting the pieces together yields a $125$M language model at baseline quality whose feed-forward units are
named set and quantifier operations throughout and whose deep attention values are readable memberships: a
model that is, over its dominant computations, legible before anyone interprets it---so that a token it
generates can be decomposed into the named units that compose to produce it. What makes this more than a
curiosity is how little it costs: the attention mechanism adds no parameters, holds quality at parity with a
conventional baseline---the best end-to-end variant level with it at the top of the parity band---and asks only that we bound what
a head detects and leave what it writes free. Legibility, long treated as something to recover after training
and at some price in accuracy, is here built in and effectively free. The remaining opacity---
shallow attention, polysemantic operands---marks the work still to do, but the object is now the right kind of
object: one built to be read, edited, and audited, rather than reconstructed after the fact.

\appendix
\section{A Trainability Boundary: Why the 75\% Design Fails}
\label{sec:trainability}

Table~\ref{tab:bench-attn} is missing a row. The membership family has a $75\%$ variant---nine bounded heads
and three conventional---and it does not train. The pattern is counterintuitive, and its cause is
instructive for anyone building bounded components into a residual stream.

\subsection{The phenomenon: non-monotonic in the bounded fraction}
The failure is \emph{non-monotonic}. \mvh{}-50 (six bounded, six conventional) trains; \mvh{}-100 (all
twelve bounded) trains; but \mvh{}-75 (nine bounded, three conventional) diverges---and it does so
reproducibly at two depths, at $4$ layers and again at $12$. Figure~\ref{fig:divergence}(a) shows the $12$-layer
run: it descends cleanly for roughly $16$k steps---tracking the healthy runs---then turns over and climbs
monotonically in both running and instantaneous perplexity until the loss-based guard aborts it at step
$30{,}250$. Depth did not fix it; it bought about a thousand steps. ``More bounding is worse'' would be
monotonic, and it is not, so the cause cannot simply be the amount of bounding.

\subsection{The cause: a starved unbounded minority}
The clue is the non-monotonicity itself. Bounded heads cap their values in $[0,1]$; conventional heads are
unbounded. They write the same residual stream, so any high-dynamic-range content the bounded heads
\emph{cannot} represent---rare tokens, large-magnitude directions---must route through the conventional
heads. At $50\%$ there are six unbounded heads to absorb that load; at $100\%$ there is no unbounded content
at all and the system self-balances. At $75\%$ there are only \emph{three} unbounded heads carrying
everything nine bounded heads cannot---overdriven, and with no bound to stop them. The danger is not the
bounded majority; it is the \emph{starved unbounded minority}.

\begin{figure}[t]
\centering
\includegraphics[width=0.94\textwidth]{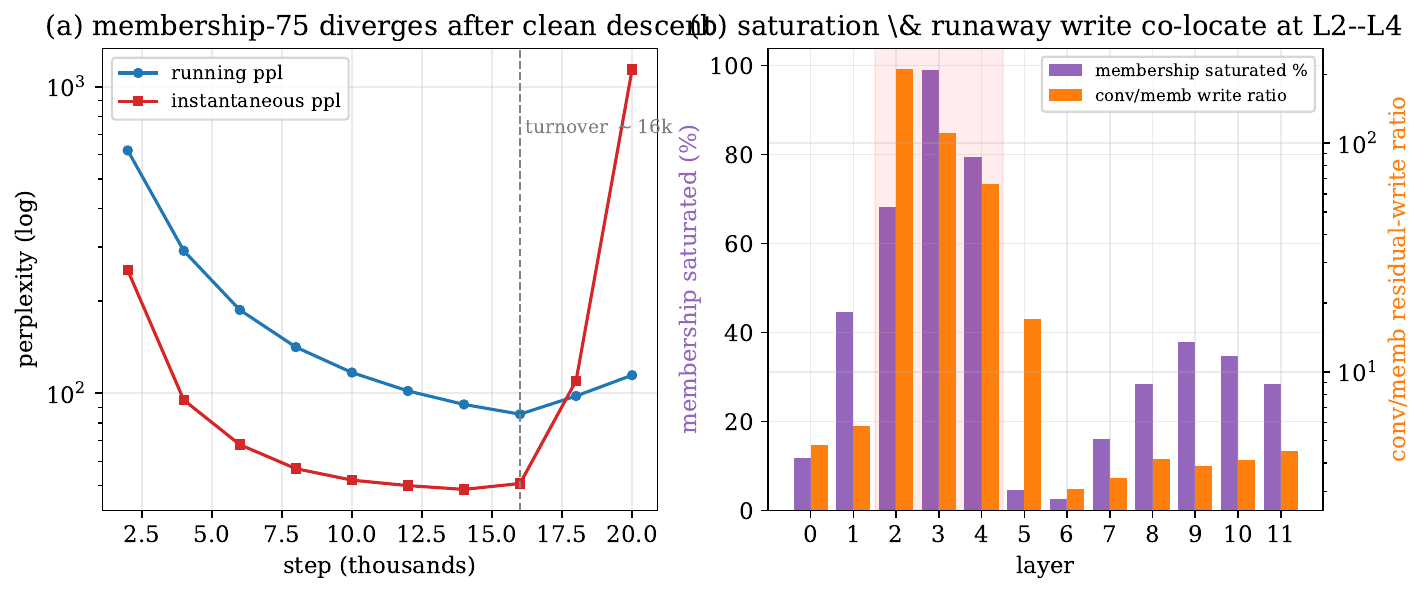}
\caption{Why \mvh{}-75 fails. (a) The run descends cleanly for $\sim\!16$k steps, then diverges in both
running and instantaneous perplexity (log scale) until aborted. (b) At the abort checkpoint, the fraction of
bounded memberships that are saturated (left axis, purple) and the ratio of conventional-to-bounded per-head
residual write (right axis, log, orange) spike together at layers L2--L4: the bounded heads freeze and the
three unbounded heads are driven $100$--$200\times$ harder to compensate.}
\label{fig:divergence}
\end{figure}

We confirm this on the saved divergence checkpoint with two reads. First, static: per-head output-projection
column norms are systematically larger for the three conventional heads than the nine bounded ones
($1.34\times$ on average, up to $2\times$ in mid-network)---the minority is overweighted on the write side.
Second, and decisively, a forward read at the abort point measures, per layer, the fraction of bounded
memberships that are \emph{saturated} (pinned within $0.02$ of a rail, where the sigmoid gradient vanishes)
and the per-head \emph{residual-write magnitude}. Figure~\ref{fig:divergence}(b) shows the two together, and
they tell one story, co-located at layers L2--L4:
\begin{itemize}
\item The bounded heads \emph{saturate} at L2--L4---$68\%$, $99\%$, $79\%$ of channels pinned to $0/1$. A
saturated sigmoid has near-zero gradient, so those heads are frozen; their per-head write collapses to
$2$--$5$.
\item The three conventional heads \emph{explode} at exactly those layers to compensate---per-head residual
writes of $441$ and $523$ at L2--L3 against $\sim\!3$ for the bounded heads, a $100$--$200\times$ ratio. A
magnitude-$500$ vector added to the residual stream is the divergence.
\item Where the bounded heads stay healthy (L6--L11, $3$--$38\%$ saturated), the conventional writes are
near-normal ($3$--$5\times$). The pathology is a \emph{mid-network} phenomenon.
\end{itemize}
The mechanism is a positive-feedback runaway: as the conventional heads take over, the residual grows, the
bounded pre-activations grow, their sigmoids saturate, the bounded heads freeze, and still more load falls on
the three unbounded heads, whose output projection grows until a step overshoots. Notably the bounded heads'
\emph{weights} are intact---this is saturation from blown-up activations, not weights driven to zero by the
sparsity pressure.

\subsection{True divergence versus a transient spike}
It is worth separating what ``fails'' means here from a benign phenomenon the bounded operators produce that is
easily mistaken for it. The \mvh{}-75 failure is a \emph{true} divergence: the \emph{cumulative} perplexity
climbs monotonically (from $\sim\!85$ at step $16$k to $230$ at the abort), the instantaneous perplexity is
\emph{sustained} in the hundreds-to-thousands, and the residual scale $\gamma$ collapses throughout
($57\!\to\!42$)---a positive-feedback runaway that a looser abort threshold only postpones, never averts.

Distinct from this, healthy bounded-operator runs throw \emph{isolated} instantaneous-perplexity spikes while
their cumulative perplexity stays flat. The operators spend most of their mass at the rails---crispness
pressure pins them there explicitly, sparsity pins them off---so they behave as near-binary gates, and on an
occasional batch a handful flip and the single-window perplexity jumps to the hundreds or beyond before
returning. In our converged runs (final perplexity $\approx\!31$), $1$--$2\%$ of logging windows exceed
instantaneous perplexity $100$ with isolated maxima near $6{,}000$, none of which move the running average.
These are not divergences: \bvh{}-75 crispness-alone, for instance, was aborted by a tight
instantaneous-perplexity guard at step $261{,}750$---$96\%$ of the way through training and sitting at a flat
perplexity of $31.7$---by a \emph{single} window that touched $705$, and it trains to completion once the guard
is loosened. The reliable divergence signal is therefore a rising \emph{cumulative} perplexity (with $\gamma$
collapse), not an instantaneous spike; a guard keyed tightly to the instantaneous value false-positives on the
binarized operators, so we set it loosely. The only genuine divergences we observe are \mvh{}-75 (at every
depth) and the fully-Boolean feed-forward layer under both pressures---both showing the sustained, cumulative
signature above; several runs first \emph{recorded} as failures were tight-guard false alarms that train
cleanly.

\subsection{Scope, caveat, and the fix}
Two controls sharpen the claim. First, the \emph{Boolean} family at the same $9\!+\!3$ split (\bvh{}-75) trains
fine (Table~\ref{tab:bench-attn}), which isolates the pathology to the sparsity-penalized \emph{membership}
family's residual-load asymmetry rather than to bounded value heads in general---the only difference between
them being a bounded sigmoid value versus a bounded \emph{multiplicative} value. Second, the limit of this
read: our forward read is a single snapshot taken $\sim\!14$k steps \emph{into} divergence, when saturation and the
runaway write are already mutually reinforcing.

This is the same class of instability as the Boolean-fraction-dependent trainability horizon of the legible
FFN \citep{oskin2026ncffn}: late-onset, abrupt, driven by saturating bounded units composing into the
residual stream. It adds a mechanistic wrinkle. There, a parallel linear ``highway'' only \emph{delayed}
divergence; here the conventional heads \emph{are} a linear highway, and they behave exactly so---three of
them delay the failure by $\sim\!16$k steps but cannot carry the load, while six (or zero) are safe. The lesson
is precise: a highway that is \emph{too narrow} is worse than none, because it draws load it cannot carry. The
practical fix follows directly---do not run a small unbounded minority: keep conventional heads at zero (the
$100\%$ design) or at least at an even split, or give the bounded majority a magnitude escape valve so the
minority is not the sole sink.

\bibliographystyle{plainnat}
\bibliography{refs}

@article{oskin2026ncffn,
  title={Explicit Fuzzy Logic in the Feed-Forward Layer: Self-Forgetting Quantifiers Discover Legible Grammatical-Licensing Detectors},
  author={Oskin, Mark},
  journal={arXiv preprint arXiv:2606.31845},
  year={2026}
}

@inproceedings{vaswani2017attention,
  title={Attention Is All You Need},
  author={Vaswani, Ashish and Shazeer, Noam and Parmar, Niki and Uszkoreit, Jakob and Jones, Llion and Gomez, Aidan N. and Kaiser, Lukasz and Polosukhin, Illia},
  booktitle={Advances in Neural Information Processing Systems (NeurIPS)},
  year={2017}
}

@article{elhage2021framework,
  title={A Mathematical Framework for Transformer Circuits},
  author={Elhage, Nelson and Nanda, Neel and Olsson, Catherine and Henighan, Tom and Joseph, Nicholas and Mann, Ben and others},
  journal={Transformer Circuits Thread},
  year={2021}
}

@misc{nostalgebraist2020logitlens,
  title={Interpreting {GPT}: The Logit Lens},
  author={{nostalgebraist}},
  howpublished={LessWrong},
  year={2020},
  note={\url{https://www.lesswrong.com/posts/AcKRB8wDpdaN6v6ru/interpreting-gpt-the-logit-lens}}
}

@article{belrose2023tunedlens,
  title={Eliciting Latent Predictions from Transformers with the Tuned Lens},
  author={Belrose, Nora and Furman, Zach and Smith, Logan and Halawi, Danny and Ostrovsky, Igor and McKinney, Lev and Biderman, Stella and Steinhardt, Jacob},
  journal={arXiv preprint arXiv:2303.08112},
  year={2023}
}

@inproceedings{geva2021kv,
  title={Transformer Feed-Forward Layers Are Key-Value Memories},
  author={Geva, Mor and Schuster, Roei and Berant, Jonathan and Levy, Omer},
  booktitle={Empirical Methods in Natural Language Processing (EMNLP)},
  year={2021}
}

@inproceedings{geva2022promote,
  title={Transformer Feed-Forward Layers Build Predictions by Promoting Concepts in the Vocabulary Space},
  author={Geva, Mor and Caciularu, Avi and Wang, Kevin Ro and Goldberg, Yoav},
  booktitle={Empirical Methods in Natural Language Processing (EMNLP)},
  year={2022}
}

@inproceedings{dar2023analyzing,
  title={Analyzing Transformers in Embedding Space},
  author={Dar, Guy and Geva, Mor and Gupta, Ankit and Berant, Jonathan},
  booktitle={Association for Computational Linguistics (ACL)},
  year={2023}
}

@article{sakarvadia2023attnlens,
  title={Attention Lens: A Tool for Mechanistically Interpreting the Attention Head Information Retrieval Mechanism},
  author={Sakarvadia, Mansi and Ajith, Aswathy and Khan, Arham and Grzenda, Daniel and Hudson, Nathaniel and Bauer, Andr{\'e} and Chard, Kyle and Foster, Ian},
  journal={arXiv preprint arXiv:2310.16270},
  year={2023}
}

@article{kissane2024attnsae,
  title={Interpreting Attention Layer Outputs with Sparse Autoencoders},
  author={Kissane, Connor and Krzyzanowski, Robert and Bloom, Joseph Isaac and Conmy, Arthur and Nanda, Neel},
  journal={arXiv preprint arXiv:2406.17759},
  year={2024}
}

@article{sun2025denselatents,
  title={Dense {SAE} Latents Are Features, Not Bugs},
  author={Sun, Xiaoqing and Stolfo, Alessandro and Engels, Joshua and Wu, Ben and Rajamanoharan, Senthooran and Sachan, Mrinmaya and Tegmark, Max},
  journal={arXiv preprint arXiv:2506.15679},
  year={2025}
}

@article{elhage2022superposition,
  title={Toy Models of Superposition},
  author={Elhage, Nelson and Hume, Tristan and Olsson, Catherine and Schiefer, Nicholas and Henighan, Tom and others},
  journal={Transformer Circuits Thread},
  year={2022}
}

@article{bricken2023monosemanticity,
  title={Towards Monosemanticity: Decomposing Language Models with Dictionary Learning},
  author={Bricken, Trenton and Templeton, Adly and Batson, Joshua and Chen, Brian and Jermyn, Adam and others},
  journal={Transformer Circuits Thread},
  year={2023}
}

@inproceedings{cunningham2023sae,
  title={Sparse Autoencoders Find Highly Interpretable Features in Language Models},
  author={Cunningham, Hoagy and Ewart, Aidan and Riggs, Logan and Huben, Robert and Sharkey, Lee},
  booktitle={International Conference on Learning Representations (ICLR)},
  year={2024}
}

@article{bills2023autointerp,
  title={Language Models Can Explain Neurons in Language Models},
  author={Bills, Steven and Cammarata, Nick and Mossing, Dan and Tillman, Henk and Gao, Leo and Goh, Gabriel and Sutskever, Ilya and Leike, Jan and Wu, Jeff and Saunders, William},
  journal={OpenAI},
  year={2023}
}

@article{elhage2022solu,
  title={Softmax Linear Units},
  author={Elhage, Nelson and Hume, Tristan and Olsson, Catherine and Nanda, Neel and Henighan, Tom and others},
  journal={Transformer Circuits Thread},
  year={2022}
}

@article{tamkin2023codebook,
  title={Codebook Features: Sparse and Discrete Interpretability for Neural Networks},
  author={Tamkin, Alex and Taufeeque, Mohammad and Goodman, Noah D.},
  journal={arXiv preprint arXiv:2310.17230},
  year={2023}
}

@inproceedings{rigotti2022concept,
  title={Attention-Based Interpretability with Concept Transformers},
  author={Rigotti, Mattia and Miksovic, Christoph and Giurgiu, Ioana and Gschwind, Thomas and Scotton, Paolo},
  booktitle={International Conference on Learning Representations (ICLR)},
  year={2022}
}

@article{pearce2024bilinear,
  title={Bilinear {MLP}s Enable Weight-Based Mechanistic Interpretability},
  author={Pearce, Michael T. and Dooms, Thomas and Rigg, Alice and Oramas, Jose and Sharkey, Lee},
  journal={International Conference on Learning Representations (ICLR)},
  note={arXiv:2410.08417},
  year={2025}
}

@article{ramapuram2024sigmoidattn,
  title={Theory, Analysis, and Best Practices for Sigmoid Self-Attention},
  author={Ramapuram, Jason and Danieli, Federico and Dhekane, Eeshan and Weers, Floris and Busbridge, Dan and Ablin, Pierre and Likhomanenko, Tatiana and Digani, Jagrit and Gu, Zijin and Shidani, Amitis and Webb, Russ},
  journal={arXiv preprint arXiv:2409.04431},
  year={2024}
}

@article{rahmanzadehgervi2024tab,
  title={{TAB}: Transformer Attention Bottlenecks Enable User Intervention and Debugging in Vision-Language Models},
  author={Rahmanzadehgervi, Pooyan and Nguyen, Hung Huy and Liu, Rosanne and Mai, Long and Nguyen, Anh Totti},
  journal={arXiv preprint arXiv:2412.18675},
  year={2024}
}

@article{vankrieken2022fuzzy,
  title={Analyzing Differentiable Fuzzy Logic Operators},
  author={van Krieken, Emile and Acar, Erman and van Harmelen, Frank},
  journal={Artificial Intelligence},
  volume={302},
  note={arXiv:2002.06100},
  year={2022}
}

@article{badreddine2022ltn,
  title={Logic Tensor Networks},
  author={Badreddine, Samy and d'Avila Garcez, Artur and Serafini, Luciano and Spranger, Michael},
  journal={Artificial Intelligence},
  volume={303},
  year={2022}
}

@inproceedings{petersen2022difflogic,
  title={Deep Differentiable Logic Gate Networks},
  author={Petersen, Felix and Borgelt, Christian and Kuehne, Hilde and Deussen, Oliver},
  booktitle={Advances in Neural Information Processing Systems (NeurIPS)},
  year={2022}
}

@article{shazeer2020glu,
  title={{GLU} Variants Improve Transformer},
  author={Shazeer, Noam},
  journal={arXiv preprint arXiv:2002.05202},
  year={2020}
}

@article{dauphin2017glu,
  title={Language Modeling with Gated Convolutional Networks},
  author={Dauphin, Yann N. and Fan, Angela and Auli, Michael and Grangier, David},
  journal={International Conference on Machine Learning (ICML)},
  year={2017}
}

@article{smolensky1990tpr,
  title={Tensor Product Variable Binding and the Representation of Symbolic Structures in Connectionist Systems},
  author={Smolensky, Paul},
  journal={Artificial Intelligence},
  volume={46},
  number={1--2},
  pages={159--216},
  year={1990}
}

@article{gers2000forget,
  title={Learning to Forget: Continual Prediction with {LSTM}},
  author={Gers, Felix A. and Schmidhuber, J{\"u}rgen and Cummins, Fred},
  journal={Neural Computation},
  volume={12},
  number={10},
  pages={2451--2471},
  year={2000}
}

@article{wortsman2023smallscale,
  title={Small-Scale Proxies for Large-Scale Transformer Training Instabilities},
  author={Wortsman, Mitchell and Liu, Peter J. and Xiao, Lechao and Everett, Katie and Alemi, Alex and Adlam, Ben and Co-Reyes, John D. and Gur, Izzeddin and Kumar, Abhishek and Novak, Roman and others},
  journal={arXiv preprint arXiv:2309.14322},
  year={2023}
}

@article{warstadt2020blimp,
  title={{BLiMP}: The Benchmark of Linguistic Minimal Pairs for English},
  author={Warstadt, Alex and Parrish, Alicia and Liu, Haokun and Mohananey, Anhad and Peng, Wei and Wang, Sheng-Fu and Bowman, Samuel R.},
  journal={Transactions of the Association for Computational Linguistics (TACL)},
  year={2020}
}

@inproceedings{paperno2016lambada,
  title={The {LAMBADA} Dataset: Word Prediction Requiring a Broad Discourse Context},
  author={Paperno, Denis and Kruszewski, Germ{\'a}n and Lazaridou, Angeliki and Pham, Quan Ngoc and Bernardi, Raffaella and Pezzelle, Sandro and Baroni, Marco and Boleda, Gemma and Fern{\'a}ndez, Raquel},
  booktitle={Association for Computational Linguistics (ACL)},
  year={2016}
}

@article{sakaguchi2020winogrande,
  title={{WinoGrande}: An Adversarial Winograd Schema Challenge at Scale},
  author={Sakaguchi, Keisuke and Le Bras, Ronan and Bhagavatula, Chandra and Choi, Yejin},
  journal={AAAI Conference on Artificial Intelligence},
  year={2020}
}

@article{clark2018arc,
  title={Think You Have Solved Question Answering? Try {ARC}, the {AI2} Reasoning Challenge},
  author={Clark, Peter and Cowhey, Isaac and Etzioni, Oren and Khot, Tushar and Sabharwal, Ashish and Schoenick, Carissa and Tafjord, Oyvind},
  journal={arXiv preprint arXiv:1803.05457},
  year={2018}
}

@misc{gao2023lmeval,
  title={A Framework for Few-Shot Language Model Evaluation},
  author={Gao, Leo and Tow, Jonathan and Biderman, Stella and Black, Sid and others},
  howpublished={EleutherAI lm-evaluation-harness, Zenodo, \url{https://doi.org/10.5281/zenodo.5371628}},
  year={2021}
}

\end{document}